\documentclass{article}

\usepackage{arxiv}

\usepackage[utf8]{inputenc} 
\usepackage[T1]{fontenc}    
\usepackage{hyperref}       
\usepackage{url}            
\usepackage{booktabs}       
\usepackage{amsfonts}       
\usepackage{nicefrac}       
\usepackage{microtype}      
\usepackage{lipsum}
\usepackage{graphicx}
\usepackage{siunitx}
\usepackage[acronym]{glossaries}
\graphicspath{ {./images/} }

\newacronym{rsnn}{RSNN}{Recurrent Spiking Neural Network}
\newacronym{snn}{SNN}{Spiking Neural Network}
\newacronym{stdp}{STDP}{spike timing-dependent plasticity}
\newacronym{cr}{CR}{Correct Rate}
\newacronym{asic}{ASIC}{Application Specific Integrated Circuit}
\newacronym{camk}{CaMK}{calcium/calmodulin-dependent protein kinase}
\newacronym{bcall}{BiCaLL}{Bistable Calcium-based Local Learning}
\newacronym{sfnn}{sFNN}{Spiking Feedforward Network}
\newacronym{clp}{CLP}{Coding-level Problem}
\newacronym{srdp}{SRDP}{spike-rate-dependent plasticity}
\newacronym{hd}{HD}{Hamming Distance}
\newacronym{adex}{AdEx}{Adaptive Exponential Integrate-and-fire}
\newacronym{lif}{LIF}{Leaky Integrate-and-fire}
\newacronym{ion}{ION}{Inferior Olivary Nucleus}
\newacronym{ltp}{LTP}{long-term potentiation}
\newacronym{egta}{EGTA}{ethylene glycol tetra-acetic acid}
\newacronym{epsp}{EPSP}{excitatory post-synaptic potential}
\newacronym{ipsp}{IPSP}{inhibitory post-synaptic potential}
\newacronym{mr}{MR}{maximal rate}
\newacronym{ar}{AR}{average rate}
\newacronym{ltd}{LTD}{long-term depression}
\newacronym{pf}{PF}{parallel fibre}
\newacronym{pc}{PC}{Purkinje cells}
\newacronym{nmda}{NMDA}{N-methyl-D-aspartate}
\newacronym{cf}{CF}{climbing fiber}
\newacronym{pdf}{PDF}{probability density function}

\title{Learning in Spiking Neural Networks with a Calcium-based Hebbian Rule for Spike-timing-dependent Plasticity\thanks{This project has received funding from the EU’s Horizon 2020 programme under the Marie Skłodowska-Curie grant agreement No 861153.}}

\author{
 Willian Soares Girão, Nicoletta Risi, Elisabetta Chicca \\
  Bio-Inspired Circuits and Systems (BICS) Lab\\
  Zernike Institute for Advanced Materials\\
  Groningen Cognitive Systems and Materials Center\\
  University of Groningen, Netherlands \\
  \texttt{\{w.soares.girao, n.risi,e.chicca\}@rug.nl} \\
}

\begin{document}
\maketitle
\begin{abstract}
Understanding how biological neural networks are shaped via local plasticity mechanisms can lead to energy-efficient and self-adaptive information processing systems, which promises to mitigate some of the
current roadblocks in edge computing systems. While biology makes use of spikes to seamless use both spike timing and mean firing rate to modulate synaptic strength, most models focus on one of the two. In this work, we present a Hebbian local learning rule that models synaptic modification as a function of calcium traces tracking neuronal activity. We show how the rule reproduces results from spike time and spike rate protocols from neuroscientific studies. Moreover, we use the model to train spiking neural networks on MNIST digit recognition to show and explain what sort of mechanisms are needed to learn real-world patterns. We show how our model is sensitive to correlated spiking activity and how this enables it to modulate the learning rate of the network without altering the mean firing rate of the neurons nor the hyparameters of the learning rule. To the best of our knowledge, this is the first work that showcases how spike timing and rate can be complementary in their role of shaping the connectivity of spiking neural networks.
\end{abstract}


\section{Introduction}
Synaptic plasticity, defined as the ability of synapses to alter their strength, is a fundamental process in biological systems that underlies learning and memory. The Hebbian learning model is one of the earliest and most influential theories of synaptic plasticity. Over seven decades ago, Donald O. Hebb laid the foundation for what would have become known as Hebbian learning by proposing that when two neurons are repeatedly and consistently activated together, the synapse connecting them is strengthened. In his seminal work~\cite{Hebb_1999} he stated ``\textit{let us assume that the persistence or repetition of a reverberatory activity (or `trace') tends to induce lasting cellular changes that add to its stability. \ldots When an axon of cell A is near enough to excite a cell B and repeatedly or persistently takes part in firing it, some growth process or metabolic change takes place in one or both cells such that A’s efficiency, as one of the cells firing B, is increased}''.

In order to gain insight into the underlying mechanisms of Hebbian learning, it is essential to develop a computational model that captures the biochemical processes that occur at the synapse. Nevertheless, the complexity of the model should be limited to allow efficient software simulations and hardware emulations. This requires a reduction in the complexity of chemical signaling into a set of fundamental components at high level of abstraction that can be effectively represented in a computational framework. For example, several experimental studies~\cite{Baudry_Lynch1979, Lynch_etal1983, Turner_etal1982, Malenka_etal1988, Ekerot_Kano1985, Sakurai1990, Brocher_etal1992} have identified calcium (Ca\textsuperscript{2+}) ions as key players in mediating synaptic plasticity. These studies have demonstrated that calcium influx serves as a trigger for downstream signaling pathways that lead to changes in synaptic strength. The modeling of calcium dynamics as exponentially decaying variables~\cite{Pfister_Gerstner2006, Fusi_etal2000} offers a straightforward approach to capturing the temporal aspects of calcium signaling at the synapse. 

Mechanistic models of synaptic plasticity are essential for elucidating the mechanisms underlying neural system learning and adaptation. Such models can also inform the development of bio-inspired algorithms for machine learning~\cite{Diehl_Cook2015, Kheradpisheh_etal2018}. Furthermore, these models are inherently compatible with the constraints of neuromorphic analog hardware, which aims to emulate the functionality of biological neural systems in hardware. The use of traces as a readout mechanism for timed events for computing weight changes in spiking neurons provides a biologically plausible and computationally efficient model of synaptic plasticity that is well-suited for implementation in neuromorphic hardware.

In the field of artificial intelligence and machine learning, there is a significant disparity between the learning mechanisms employed and those observed in humans and animals. Although artificial systems have made considerable progress in solving complex tasks, they frequently depend on models and architectures that are fundamentally different from those observed in biological organisms. 
Furthermore, they rely on large amounts of data and have distinct operational phases that artificially separate the learning process from the retrieval of information.
In contrast, learning in biological systems is continuous (always on), operating and generalizing on the basis of very little data.

In this paper, we present a local learning rule for spiking neurons that encodes pre- and post-synaptic activity via calcium traces and computes weight changes based on these. Our work emphasises the necessity of abstracting and modeling the intricate biochemical processes underlying local learning in synaptic plasticity. This is crucial for the development of bio-inspired learning systems that operates within the same constraints as biological ones.
By modeling the principles of synaptic plasticity and neuronal activity observed in biological systems, we pave the way for the development of artificial systems capable of learning in a manner more closely aligned with biological processes. 

To the best of our knowledge, this is the first work that shows the interplay between spike timing and mean firing rate within a network of spiking neurons. More precisely, we show how, without modifying neither network nor the learning rule hyperparameters, the timing of spikes can be used to modulate the rate of change of synaptic couplings.

The code related to the spiking networks simulations in this work can be found in this GitHub repository: https://github.com/Willian-Girao/bcall.

\section{Plasticity in Neural Systems}

It is widely accepted that activity-dependent synaptic plasticity  is the fundamental process underlying learning and memory. 
Synaptic modifications occur when there is a concurrent activition of pre- and post-synaptic neurons. As a result, they are driven by patterns of neuronal firing (e.g. reflecting sensory stimulation or ongoing brain activity), and they occur over a wide range of timescales. While short-term synaptic plasticity occurs at timescales ranging from tens of milliseconds to a few minutes~\cite{Zucker_Regehr2002}, long-term synaptic plasticity facilitates changes that can persist for hours or even days~\cite{Citri_Malenka2008}.

It is now evident that synaptic modification may occur at either side of the synapse. Post-synaptic modifications typically entail alteration in the number or properties of post-synaptic receptors, whereas pre-synaptic plasticity encompasses an increase or decrease in the release of neurotransmitters~\cite{Yang_Calakos2013}. A growing body of experimental research has investigated different methods for evoking synaptic changes, the role of local variables (e.g. post-synaptic voltage, pre-synaptic spike) and the resulting synaptic dynamics~\cite{Wittenberg_Wang2006, Bi_Poo1998, Sjostrom_etal2001, Markram_etal1997, Malenka_Bear2004}. 
Furthermore, evidence indicates that distinct synapses and brain regions necessitate disparate levels of activity to facilitate synaptic plasticity~\cite{Abbott_Nelson2000}. The biochemical and physiological processes that underpin the induction and expression of synaptic plasticity represent a significant area of investigation~\cite{Gulyaeva2017}.

\subsection{The Role of Calcium}

The role of calcium in synaptic plasticity has been demonstrated in studies conducted in the late 70's on hippocampal neurons. An early study~\cite{Baudry_Lynch1979} established that even low levels of Ca\textsuperscript{2+} concentration were effective in enhancing glutamate (a major excitatory neurotransmitter in the mammalian central nervous system) binding sites in purified membranes from the hippocampus. Subsequent studies were conducted to further investigate this relationship, providing evidence that \gls{ltp} induction is associated with a calcium-mediated alteration in the post-synaptic neuron. 

In~\cite{Lynch_etal1983}, the authors demonstrated that the intracellular injection of the chelator \gls{egta}, followed by a series of high-frequency stimulation trains, inhibits the development of hippocampal \gls{ltp} in CA1 neurons.
Based on this result, they proposed that buffering intracellular Ca\textsuperscript{2+} prevents the activation of the enzymatic machinery controlling post-synaptic receptors. 

The work presented in~\cite{Turner_etal1982} demonstrated that a transient elevation in extracellular Ca\textsuperscript{2+} concentration induces potentiation. By perfusing CA1 neurons in a high Ca\textsuperscript{2+} medium, the researchers demonstrated that both Schaffer collateral commissural evoked \gls{epsp} and population spikes (synchronous discharge of a neuronal population) were potentiated by 20-90\% and 50-640\%, respectively.

Building upon these works, the authors in~\cite{Malenka_etal1988} conducted a series of experiments on hippocampal pyramidal cells, two of which employed the use of nitr-5 chelator. 
In agreement with the findings presented in~\cite{Baudry_Lynch1979}, the authors demonstrated that an elevation in intracellular Ca\textsuperscript{2+} concentration through photolysis of Ca\textsuperscript{2+}-loaded nitr-5 is accountable for a considerable enhancement in the amplitude and initial slope of the \gls{epsp}. Furthermore, the role of the post-synaptic membrane potential was investigated.
A strong depolarization inhibits the influx of Ca\textsuperscript{2+} through \gls{nmda} channels, thereby preventing the enhancement of the \gls{epsp} in response to low-frequency stimulation. 
Conversely, a moderate depolarization gives rise to a pronounced potentiation.
It can therefore be concluded that potentiation is blocked either by buffering changes in intracellular Ca\textsuperscript{2+} or by retarding its influx with a large membrane depolarization.

The involvement of Ca\textsuperscript{2+} in the induction of \gls{ltd} in the cerebellum has been hypothesised based on findings demonstrating that depression of \gls{pf} responses is influenced by the membrane potential of \gls{pc}~\cite{Ekerot_Kano1985}. This suggests that an increase in Ca\textsuperscript{2+} concentration in the PC's dendrites may modulate changes in synaptic efficacy of \glspl{pf}. 
Direct evidence for this hypothesis was provided by the investigation in~\cite{Sakurai1990}, which demonstrated that when \gls{egta} is injected into \glspl{pc}, the conjunctive activation of \gls{cf} and \gls{pf} no longer produces \gls{ltd}. This suggests that the action of \gls{cf} impulses in inducing \gls{ltd} is canceled by Ca\textsuperscript{2+} chelation.

Furthermore, evidence for calcium-mediated \gls{ltd} was identified in other regions of the brain. The work presented in~\cite{Brocher_etal1992} demonstrated that \gls{ltd} induced by tetanic stimulation of afferent fibres ascending from the white matter to layer 3 cells in the rat visual cortex can be reliably blocked by intracellular injection of Ca\textsuperscript{2+} chelators, indicating that \gls{ltd} requires a minimal post-synaptic intracellular Ca\textsuperscript{2+} concentration. The authors highlight given the dependence of both \gls{ltd} and \gls{ltp} on intracellular Ca\textsuperscript{2+} concentration, it is plausible that activation conditions resulting in a substantial surge of intracellular Ca\textsuperscript{2+} may favor the occurrence of \gls{ltp}, whereas smaller increases may lead to \gls{ltd}. This is in agreement with the evidence that \gls{ltp} induction requires, in addition to a large post-synaptic depolarization, the activation of \gls{nmda} receptor-dependent Ca\textsuperscript{2+} conductances~\cite{Baudry_Lynch1979}, while \gls{ltd} can also be induced following the blockade of \gls{nmda} receptors.

The collective findings of these studies indicate that both \gls{ltp} and \gls{ltd} appear to involve calcium-mediated processes. The calcium hypothesis, as outlined by~\cite{Sjostrom_etal2008}, states that \gls{ltp} is triggered by a brief elevation in the post-synaptic calcium concentration, whereas \gls{ltd} is induced by a smaller and more prolonged increase in calcium concentration.

\subsection{Calcium-based Hebbian Learning Rule}

As the set of studies we reviewed~\cite{Baudry_Lynch1979, Lynch_etal1983, Turner_etal1982, Malenka_etal1988, Ekerot_Kano1985, Sakurai1990, Brocher_etal1992} indicates, calcium plays an important role in both \gls{ltd} and \gls{ltp} of synaptic strength. Therefore, it is unsurprising that mechanistic models (learning rules) of synaptic changes based on calcium-mediated processes have been previously proposed~\cite{Brader_etal2007,Pfister_Gerstner2006,Graupner_Brunel2007,Graupner_Brunel2010,Graupner_Brunel2012}. 
Given that the locus of expression of plasticity can be either pre-synaptic~\cite{Markram_Tsodyks1996, Sjostrom_etal2003}, post-synaptic~\cite{Lisman2003}, or both~\cite{Sjostrom_etal2007}, we present a novel mechanistic model of local Hebbian learning in which the weight update is based on two  synaptic variables (referred to as traces) modeling changes in intracellular calcium transients triggered by pre-synaptic and post-synaptic spikes. 

We refer to this mechanistic model of synaptic plasticity as the \gls{bcall} rule. This model is capable of reproducing the experimental results of both \gls{stdp} and \gls{srdp} protocols, whereby the direction and magnitude of synaptic changes can be controlled by the timing of repeated pre- and post-synaptic spike patterns.

The \gls{bcall} rule was inspired by an abstract triplet model similar to that proposed in~\cite{Pfister_Gerstner2006}, the maintenance model put forth in~\cite{Fusi_etal2000} and by previously proposed models that utilise calcium dynamics~\cite{Graupner_Brunel2007, Graupner_Brunel2010, Graupner_Brunel2012,Morrison_etal2008}. A common feature of these models and our approach is the use of exponentially decaying traces, updated in an event-based manner. This is a well-established approach for the design of biologically inspired learning rules, as evidenced by the extensive literature on the subject~\cite{Khacef_etal2023}. The design of our model was done with an analog circuit implementation~\cite{Willian_etal2024_cognigr1} in mind, making it suitable for learning neuromorphic systems.

The objective of our modeling work was to incorporate synaptic dynamics at the level of spike timing and mean firing rate. 
The calcium traces of the model are crucial for realizing the dependency of the synaptic weight on these two time scales.
From a theoretical standpoint, this model can be classified as an unsupervised learning rule. In other words, learning is an adaptation of the synapse to the statistics of activity of pre- and post-synaptic neurons. 
The synapse is able to access information pertaining to neuronal activity via the pre- and post-synaptic calcium concentrations, represented by $x_{i}$ and $x_{j}$ (calcium traces or simply traces), respectively. Henceforth, the subscript ``$i$'' shall denote a pre-synaptic neuron, while ``$j$'' shall refer to the respective post-synaptic neuron. 

In response to neuronal spiking activity, the calcium concentration increases and subsequently diffuses away. The dynamic of the calcium concentration at the pre- and post-synaptic neuron can therefore be described by the following equations.

\begin{equation} \label{eq:pre_trace_eq}
\frac{dx_{i}}{dt} = \frac{-x_{i}}{\tau_{i}} + \sum_{pre \hspace{1mm}k}a_{i}[x_{i}^{max} - x_{i}(t-\epsilon)]\delta(t-t_{k})
\end{equation}


\begin{equation} \label{eq:post_trace_eq}
\frac{dx_{j}}{dt} = \frac{-x_{j}}{\tau_{j}} + \sum_{post \hspace{1mm}k}a_{j}[x_{j}^{max} - x_{j}(t-\epsilon)]\delta(t-t_{k})
\end{equation}


Equations~\ref{eq:pre_trace_eq} and~\ref{eq:post_trace_eq} describe the evolution of the calcium traces in a pre-synaptic neuron $i$ and a post-synaptic neuron $j$. A spike in these equations is modeled by a Dirac delta function $\delta(t-t_{k})$, where $t_{k}$ represents the time when the respective neuron fires an action potential. The constant $\epsilon$ is assumed to be infinitesimally small. The time constant $\tau$ governs the decay of calcium concentration taking place in absence of spikes. 

These aforementioned trace variable $x_{n}$ are increased by an amount $a_{n}[1 - x_{n}(t-\epsilon)]$ proportional to their current value in response to a spike (with $n \in [i, j]$). The variable $a_{n}$ represents a scaling factor, specifically a trace update hyperparameter. As these traces model real physical variables that are bounded by limited resources, we add the soft-bound term $[1 - x_{n}(t-\epsilon)]$ to them. This results in the amplitude of the trace jump being multiplied by the difference between their maximum and current value. This soft-bound implies that the traces are asymptotically influenced by spiking history of the neuron. In this manner the interactions between spikes are said to be all-to-all, which contrasts to a nearest-spike interaction setting in which the traces would always be updated to a fixed value ($a_{i} = 1$, also referred to as a ``capped'' trace), thus being a representation of only the most recent spike~\cite{Khacef_etal2023}.

In the absence of activity the traces undergo exponential decay, returning to their baseline values with time constant $\tau_{n}$. The term $x_{n}(t-\epsilon)$ represents the value of a trace immediately prior to the update.

The traces $x_{i}$ and $x_{j}$ are employed for the purpose of translating the activity of connected neurons into synaptic weight updates. Such updates are triggered by both pre- and post-synaptic spikes, with each reading out the value of the trace sitting on the opposite side of the synapse. For example, a weight change caused by a pre-synaptic at time $t^{i}$ spike depends on $x_{j}(t^{i})$.

At the time a pre-synaptic spike is generated, the post-synaptic calcium trace is compared to a threshold $\theta_{j}$. If the trace is above it, the post-synaptic neuron has recently generated a spike, meaning a post-pre spike pairing is being processed by the synapse, which translates into a negative weight update of the internal weight variable $w_{hid}$. Equation~\ref{eq:post_pre_pairing_eq} implements this weight decrease as a function of a post-pre spike pairing, where $\Theta$ denotes the Heaviside step function and $c_{1}^{d}$ is a constant negative value (i.e. a weight decrease hyperparameter).

\begin{equation}\label{eq:post_pre_pairing_eq}
\frac{dw_{hid}}{dt} = \rho\sum_{pre\hspace{1mm}k} c_{1}^{d}\Theta(x_{j} - \theta_{j})\delta(t-t_{k})
\end{equation}

The $\rho$ in Eqs.~\ref{eq:post_pre_pairing_eq} and~\ref{eq:pre_post_pairing_eq} is a binary variable acting as a gating mechanism and it'll be explained later in this section.

At the time a post-synaptic spike is generated, the pre-synaptic calcium trace is read. If $x_{i} > 0$ at the time of the post-synaptic spike, a pre-synaptic spike has happened in the recent past and a pre-post spike pairing is being processed by the synapse. Different from the fixed discrete weight decrease we described previously, the weight increase shown in Eq.~\ref{eq:pre_post_pairing_eq} is a function of $x_{i}$, where $c^{p}$ is a positive constant scaling the weight increase (i.e. a weight increase hyperparameter). We impose this constraint in the model as way of optimizing future analog \gls{asic} design, given that a digital signal (depression as fixed update) is more reliably transmitted to many pre-synaptic neurons than an analog one (potentiation as a function of trace) and this further helps in reducing footprint.

\begin{equation}\label{eq:pre_post_pairing_eq}
\frac{dw_{hid}}{dt} = \rho\sum_{post\hspace{1mm}k}[x_{i}c^{p} + c^{d}_{2}\Theta(\theta_{i}-x_{i})]\delta(t-t_{k})
\end{equation}

Given that there's not threshold for a time since last pre-syanptic spike for the weight increase in Eq.~\ref{eq:pre_post_pairing_eq}, we add a second depression term that penalizes pre-post pairings with a large time gap between the spikes. If the pre-synaptic calcium trace is bellow a threshold $\theta_{i}$, the internal weight variable is reduced by a constant value $c_{2}^{d} \ll c_{1}^{d}$ (i.e. a weight decrease hyperparameter). We note that the conditions for decreasing the weight on a post-synaptic spike are both $x_{i}$ bellow $\theta_{i}$ and $x_{i} > 0$ (such that a post-synaptic neuron can't change $w_{hid}$ independent of the presynaptic neuron), we omit this second condition from Eq.~\ref{eq:pre_post_pairing_eq} for simplicity.

\begin{figure}[htbp]
\centerline{\includegraphics[width=10cm]{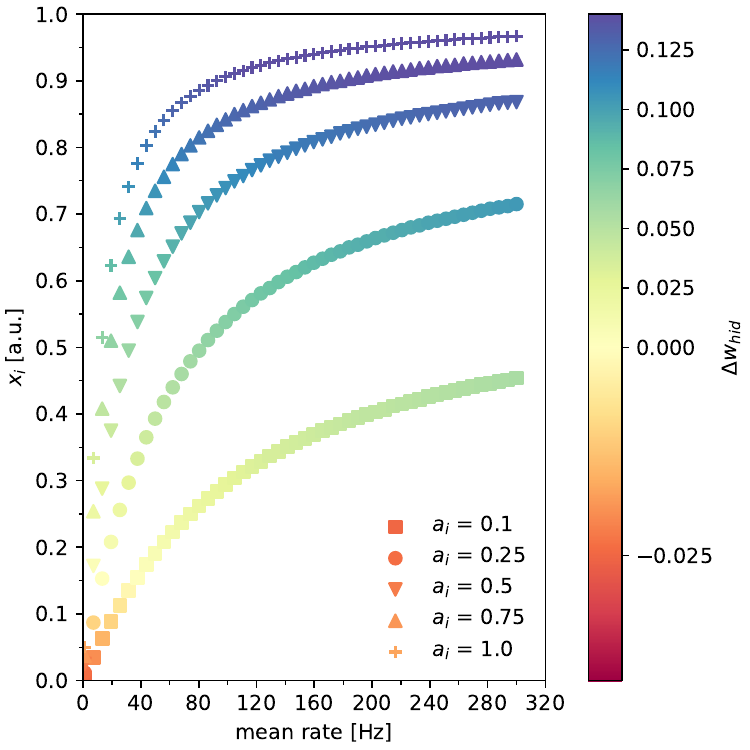}}
\caption{Mean $x_{j}$ trace value as a function of firing rate for different $a_{i}$. With a trace jump of 1 the model restricts synaptic weights to the effects of only nearest-neighbor spike pairs (i.e. jump to maximum value). The colors show how the transition point between net weight decrease and net increase shift within the mean rate spectrum: notice how for $a_{i}=1$ the net $w_{hid}$ change goes from negative to positve within a very narrow range of the mean rate whereas for $a_{i}=0.1$ this range is much wider.}
\label{fig_aj}
\end{figure}

Since potentiation is a function of the pre-synaptic trace, the transition point in the mean rate regime between net depression and net potentiation depends on the rate of integration of the variable $x_{i}$. We show this in \figurename~\ref{fig_aj} by sweeping the hyperparameter $a_{i}$ in Eq.~\ref{eq:pre_trace_eq} and mean rate at which the pre-synaptic neurons spikes, keeping the total number of spikes fixed at 20. For each data point in the plot, the weight change to be applied to the hidden variable is computed as $\Delta w_{hid} = x_{i}\times c^{p} + c^{d}_{1}$ (with this we are assuming that every potentiation event is accompanied by a depression event). With a trace update amplitude of $1$ the model restricts weight updates to the effects of only nearest-neighbor spike pairs (jump to maximum value), for values smaller than 1 the traces are integrative and encode the recent history of the neuron activity, which can be seen as an all-to-all spike interaction. What the data in \figurename~\ref{fig_aj} shows is that, depending on how spikes interact in the model (nearest-neighbor versus all-to-all), the mean rate window where net depression transitions into net potentiation changes.

In our synaptic model depression can also happen for pre-post spike pairings, as indicated by the $c_{2}^{neg}$ parameter on the right hand side of Eq.~\ref{eq:pre_post_pairing_eq}. This parameter is introduced to control the mean firing rate at which net potentiation is achieved since the parameter $c_{2}^{neg}$ can be tuned to cancel out the term $x_{pre}\times c^{pot}$ in the equation within a range of pre-synaptic activity, effectively controlling the firing frequency at which the trace of the pre-synaptic neuron accumulates enough such that $x_{pre}\times c^{pot} > c_{2}^{neg}$. Another advantage of this possibility to have depression for pre-post spike pairings is that it allows the model to produce a STDP curve observed in CA3-CA1 synapses~\cite{Wittenberg_Wang2006}.

Beyond how synaptic updates take place, in our model we also consider \textit{when} they should take place by imposing a learning window. This learning window is referred to as stop-learning condition, as in the perceptron learning rule, and it has been presented in~\cite{Brader_etal2007} as a solution to problems emerging from encoding patterns of different sizes - referred to as the \gls{clp}. The purpose behind this mechanism is to prevent weight change once the total pre-synaptic input drives the post-synaptic neuron outside a window of activity defined by two thresholds. We simplify the implementation used in~\cite{Brader_etal2007} by reducing the stop-learning thresholds from four to two. The implementation relies on having a second trace associated with post-synaptic spiking, similar to $x_{j}$ but with a longer time constant $\tau_{s}$. The weight update is stopped whenever this trace is outside a learning window defined by thresholds $\theta^{u}$ and $\theta^{l}$:

\begin{equation} \label{eq:stop_trace_eq}
\frac{dx_{s}}{dt} = \frac{-x_{s}}{\tau_{s}} + \sum_{post \hspace{1mm}k}a_{s}[x_{s}^{max} - x_{s}(t-\epsilon)]\delta(t-t_{k})
\end{equation}

\begin{equation}\label{eq:stop_trace_binary_eq}
\rho = 
\begin{cases}
  1 & \text{if $x_{s} \geq \theta_{l}$ and $x_{s} \leq \theta_{u}$} \\
  0 & \text{otherwise}
\end{cases}
\end{equation}

Equation~\ref{eq:stop_trace_eq} describes how this trace is updated during spiking of the post-synaptic neuron and how it decays in the absence of activity. Equation~\ref{eq:stop_trace_binary_eq} defines the learning window as a boolean variable controlling whether or not the updates in equations \ref{eq:post_pre_pairing_eq} and \ref{eq:pre_post_pairing_eq} are effectively applied. We will come back to this mechanism and its influence in learning in the next section.

Some studies have suggested that synapses could be similar to binary switches~\cite{Petersen_etal1998, OConnor_etal2005}  and that this would be advantageous for biological neural systems in that they would be less susceptible to noise. Similarly, in our model the synaptic weight is bistable and it incorporates a bistability term similar to the one in~\cite{Fusi_etal2000}. The weight drifts with a time constant $\tau_{w}$ either to its maximum or minimum value with slopes $\alpha$ and $\beta$, respectively, depending on whether it is above or below a threshold $\theta_{w}$. That is, even though the weight state variable $w$ is continuous the effective weight (as seen by the two connected neurons) is a binary value. This bistability is defined by Eq.~\ref{eq:bistability_eq}.

\begin{equation}\label{eq:bistability_eq}
\tau_{w}\frac{dw_{hid}}{dt} = 
\begin{cases}
  \alpha & \text{if $w_{hid} \geq \theta_{w}$} \\
  -\beta & \text{otherwise}
\end{cases}
\end{equation}

\begin{equation}\label{eq:w_eff}
w_{eff} = 
\begin{cases}
  w_{pot} & \text{if $w_{hid} \geq \theta_{w}$} \\
  w_{dep} & \text{otherwise}
\end{cases}
\end{equation}

\begin{figure*}[htbp]
\centerline{\includegraphics[width=\linewidth]{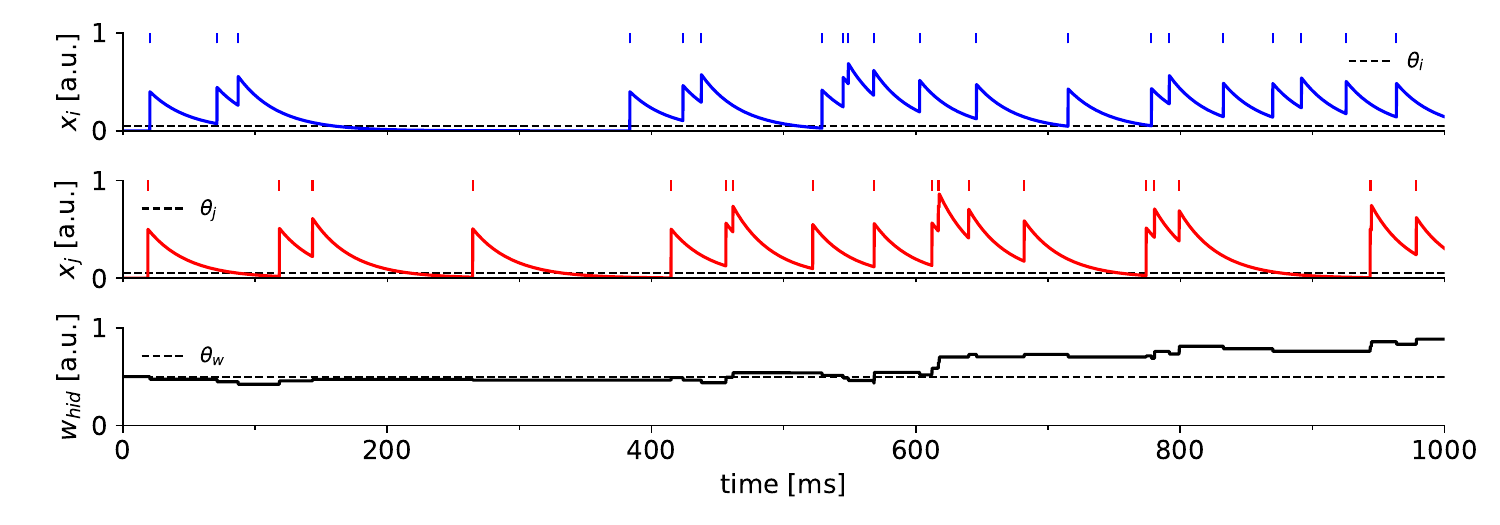}}
\caption{One second simulation of a pair of neurons connected via a plastic synapse reproducing \gls{bcall}. The three main components of the rule are shown: the pre- ($x_{i}$, blue) and post-synaptic ($x_{j}$, red) calcium traces, updated at every spike, and the trace representing the weight hidden variable ($w_{hid}$, black). Each neuron is emitting Poisson spike trains at \SI{20}{\hertz}. Hyperparameters in Table \ref{tab:hyperparams_table}.} 
\label{fig_bcal}
\end{figure*}

The traces and weight evolution during simulation of a single synapse between two neurons spiking at 15Hz and 20Hz are show in Figure \ref{fig_bcal}.

\begin{table*}
    \centering
    \begin{tabular}{c p{0.45\linewidth} c}
 Hyper-parameter & Description & Unit\\
 \midrule
         $\tau_{i}$ & Time constant of pre-synaptic calcium trace. & \SI{}{\second}\\
         $\tau_{j}$ & Time constant of the post-synaptic calcium trace. & \SI{}{\second}\\
         $\tau_{s}$ & Time constant of the stop-learning calcium trace. & \SI{}{\second}\\
         $\tau_{w}$ & Time constant of the weight bistability drift. & \SI{}{\second}\\
         $w_{eff}$ & Effective binary weight. & \SI{}{\volt}\\
\midrule
         $a_{i}$ & Amplitude of the pre-synaptic trace updated. & a.u.\\
         $a_{j}$ & Amplitude of the post-synaptic trace update. & a.u.\\
         $a_{s}$ & Amplitude of the stop-learning trace update. & a.u.\\
         $\alpha$ & Bistability's slew rate towards the UP state. & a.u.\\
         $\beta$ & Bistability's slew rate towards the DOWN state. & a.u.\\
         $\theta_{i}$ & pre-synaptic calcium trace threshold. & a.u.\\
         $\theta_{j}$ & post-synaptic calcium trace threshold. & a.u.\\
         $\theta_{u}$ & Upper stop-learning trace threshold. & a.u.\\
         $\theta_{l}$ & Lower stop-learning trace threshold. & a.u.\\
         $\theta_{w}$ & Threshold of weight bistability. & a.u.\\
         $c_{1}^{d}$ & Amplitude of weight drepression on pre-synaptic spike. & a.u.\\
         $c_{2}^{d}$ & Amplitude of weight drepression on post-synaptic spike. & a.u.\\
         $c^{p}$ & Scaling factor of potentiation amplitude on post-synaptic spike. & a.u.\\
         $x_{i}$ & Pre-synaptic calcium trace. & a.u.\\
         $x_{j}$ & Post-synaptic calcium trace. & a.u.\\
         $x_{s}$ & Stop-learning calcium trace. & a.u.\\
         $w_{hid}$ & Continuous hidden weight variable. & a.u.\\
 \midrule
    \end{tabular}
\caption{Hyperparameters of \gls{bcall}.}
\label{tab:BCaLL_variables}
\end{table*}

\subsection{Time and Frequency}

\gls{stdp} is a form of temporally asymmetric Hebbian learning where \gls{ltp} and \gls{ltd} are induced due to temporal correlations between pre- and post-synaptic spike pairings on the milliseconds time scale. The change of the synapse plotted as a function of the relative timing of pre- and post-synaptic spikes is called the STDP function and it varies between synapse types~\cite{Bi_Poo1998, Abbott_Nelson2000}. This form of plasticity was first observed in cultures of rat hippocampal neurons~\cite{Bi_Poo1998}, where the relative timing between the spike pairs determined the direction and the magnitude of synaptic changes, opening up the possibility of temporal coding schemes on the millisecond time scale.

\begin{figure*}[t]
    \centering
    {{\includegraphics[width=8cm]{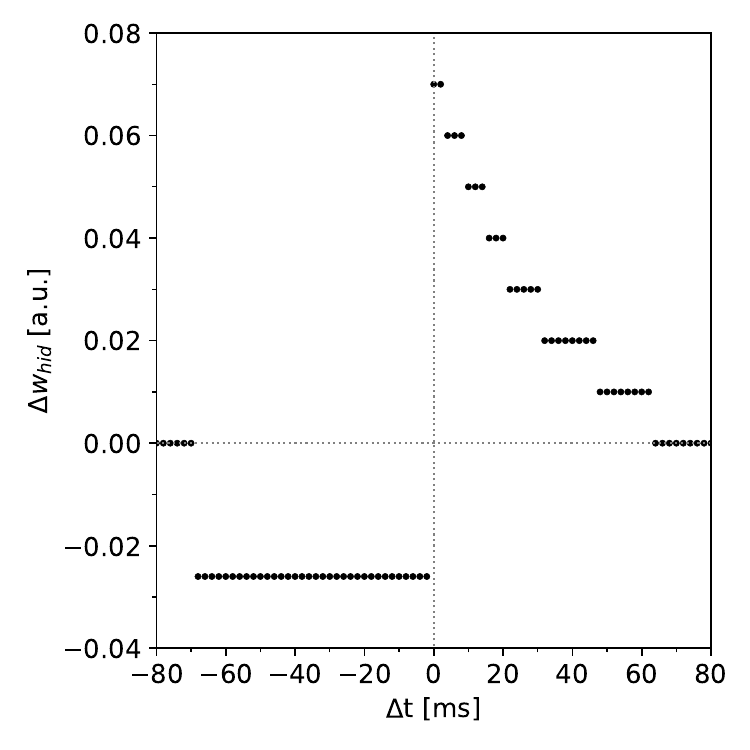}}}
    {{\includegraphics[width=8cm]{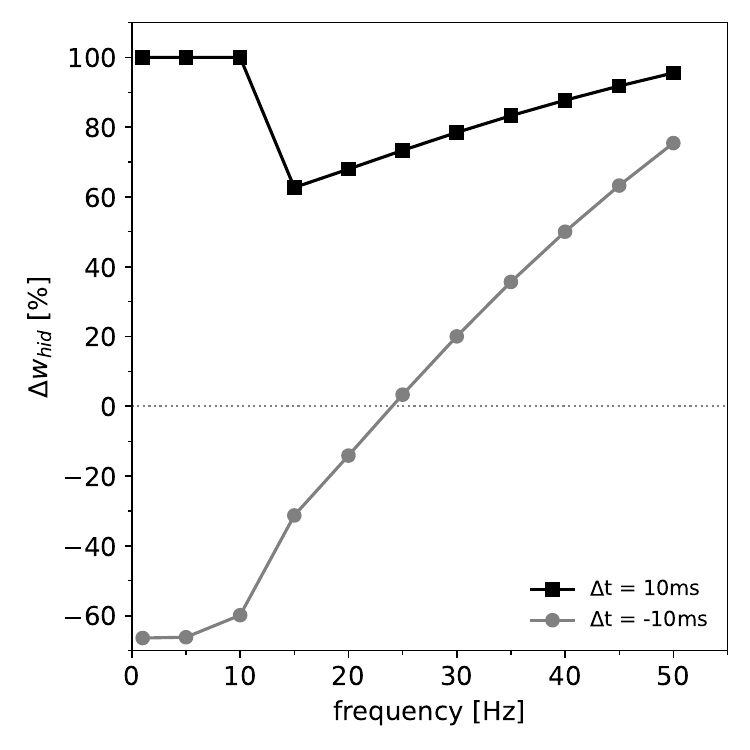}}}
    \caption{STDP (lef plot) and SRDP (right plot) curves generated by the BCaLL rule. STDP: the x-axis shows the time difference $\Delta t = t_{j} - t_{i}$ between pre- and post-synaptic spike times. The curve matches the depression for post-pre and potentiation for pre-post spike pairings; \gls{srdp}: change in synaptic weight as a function of spike pairing frequency with fixed $\Delta$t. Hyperparameters in Table \ref{tab:hyperparams_table}.}
    \label{stdp_n_srdp}
\end{figure*}

The \gls{stdp} curve in \figurename~\ref{stdp_n_srdp} (left) shows the change of synaptic weight as a function of the relative timing of pre- and post-synaptic spikes prescribed by the \gls{bcall} rule. The curve is generated by pairing single pre- and post-synaptic spikes with different $\Delta t$ from -60ms to 60ms. A noticeable difference when comparing this STDP curve with the one in~\cite{Bi_Poo1998} is that the amount of depression seen for negative time differences here is constant. This comes from a modeling decision that aims to simplify analog hardware implementations of the rule.

A different plasticity dynamic emerges if the point of view about at what time scale information is being encoded within neuronal activity changes. Experimental work~\cite{Sjostrom_etal2001} has shown that plasticity in the L5 neurons in rat visual cortex is frequency dependent, which we refer to here as \gls{srdp}. Pairings of spike pairs with fixed $-\Delta$t at low frequencies induced depression while net potentiation was achieved at high frequencies. Potentiation due to pairings with positive $\Delta$t increased with frequency, saturating around 50Hz. While other models might capture this frequency dependence by having multiple traces with different time scales~\cite{Pfister_Gerstner2006}, in \gls{bcall} this is due to the asymmetry between potentiation and depression coming from the potentiation's amplitude being a function of the pre-synaptic calcium trace (Eq.~\ref{eq:pre_post_pairing_eq}), which could be interpreted as a homeostatic mechanism. We show this plasticity dynamics in \figurename~\ref{stdp_n_srdp} (right), where a fixed number of spike-pairs (10) is simulated being emitted at increasingly higher frequencies with both positive and negative $\Delta t$.

Given that \gls{bcall} is able to reproduce spike-time and spike-rate plasticity outcomes, the former can be exploited to modulate the synaptic dynamics of the latter. Specifically, by introducing correlations in the spike pairs of two neurons connected via a synapse with this learning rule, we can bias the synaptic updates either towards more depression ($\Delta t < 0$) or potentiation ($\Delta t > 0$) events.
Temporal correlations were introduced following the algorithm proposed by~\cite{sonntag_2020}.
With $s$ being a source Poisson point process $s$ of frequency $f_s$ and $n$ spikes~\cite{Dayan_Abbott2005}, a target spike train $g$ with the frequency $f_g$ is generated as a shifted version of $s$, with $f_s \geq f_g$. The steps to generate $g$ are as follows:
\begin{enumerate}
    \item 
    $n$ uniformly distributed time shifts $L_k$, for $k \in [0, ..., n]$, are sampled from the uniform distribution $U (-\Delta t, +\Delta t)$ ,
    with $\Delta t=\gamma/f_s$.
    \item 
    The target spike train is generated as:
        $$g_k = (s_k + L_k)( rand () < P )$$
    with $P = f_g / f_s$ and for $k \in [0, ..., n]$.
    \item 
    Spikes in the target spike train $g$ with $t_i - t_{i-1} < dt$ are removed.
\end{enumerate}
%

\begin{figure*}[t]
    \centering
    {{\includegraphics[width=5.4cm]{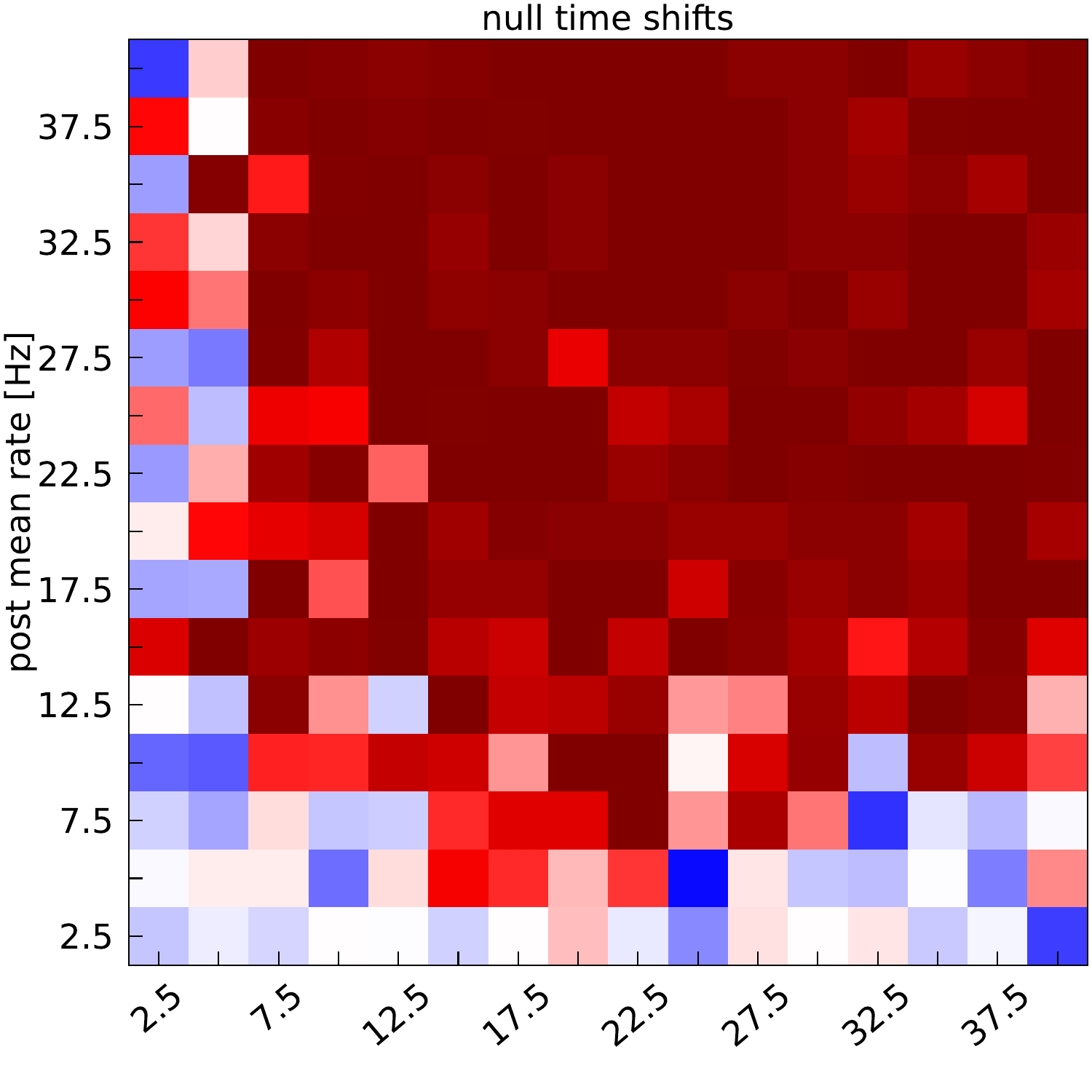}}}
    {{\includegraphics[width=5.4cm]{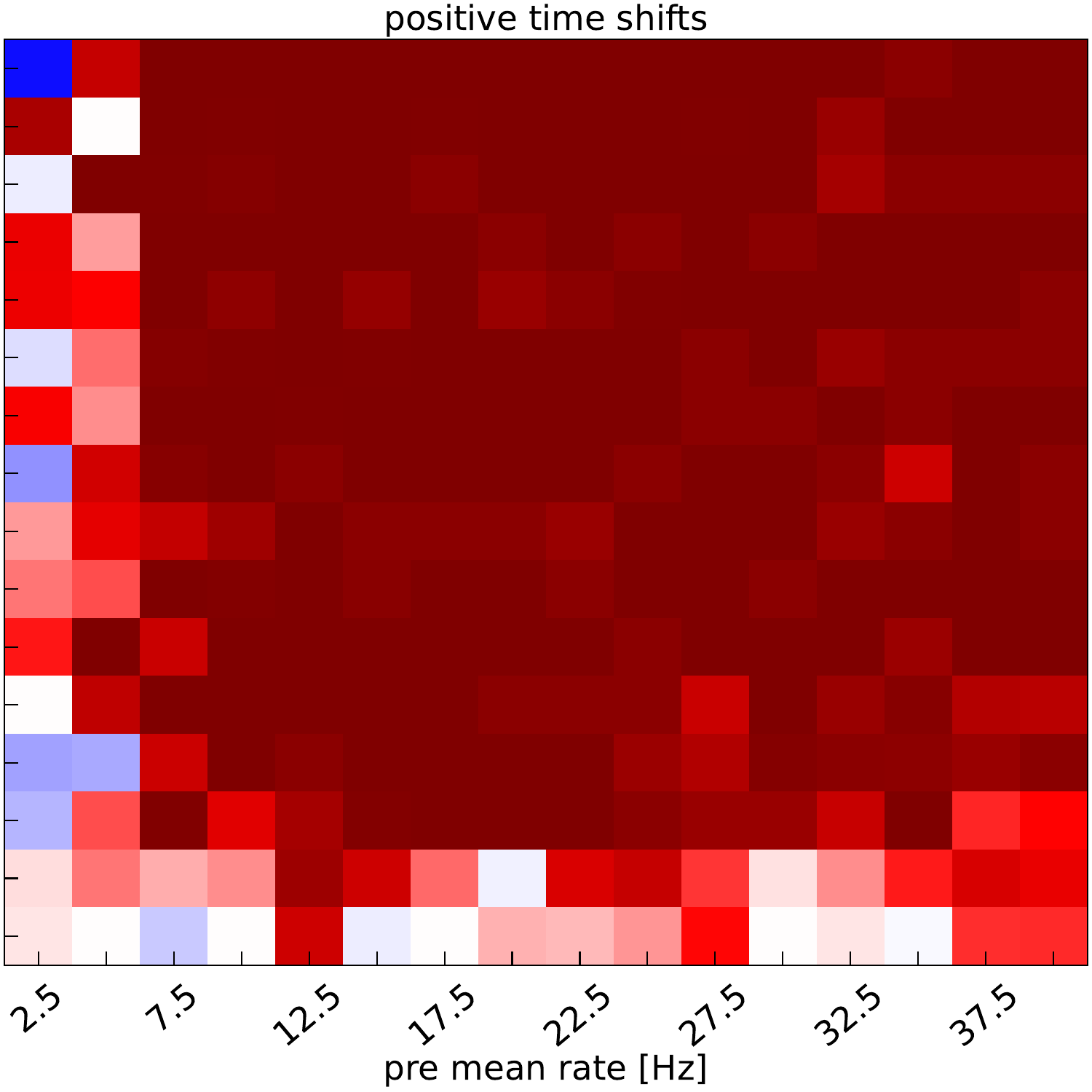}}}
    {{\includegraphics[width=5.4cm]{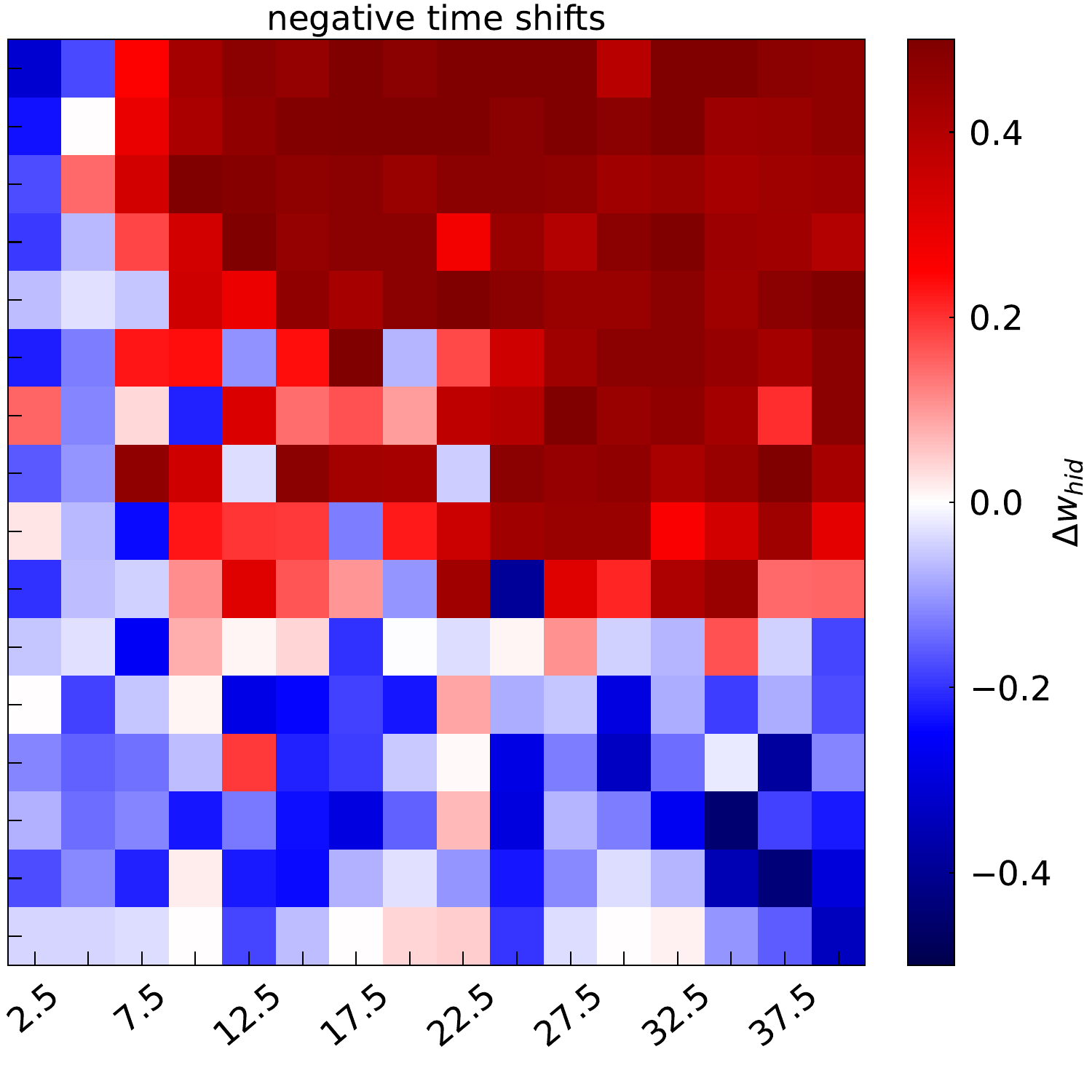}}}
    \caption{Weight change as a function of mean rate pairing of pre and post neurons. While in the leftmost heat-map each neuron in the pair emits independent Poisson spike trains, in the middle and right most heat-maps positive and negative, respectively, time shifts are introduced in the spike pairs to bias them towards either post-pre or pre-post pairings. Hyperparameters in Table \ref{tab:hyperparams_table}.}
    \label{meanrate_heatmap}
\end{figure*}

The heat-maps in \figurename~\ref{meanrate_heatmap} show how the weight state variable $w_{hid}$ changes for different pairings of mean rate activity of pre- and post-synaptic neurons, for different values of $\Delta t$ (with $\gamma=0.75$). At the beginning of the simulation, the weight variable $w_{hid}$ is set to $0.5$ and, after two seconds of stimulation of the neurons, its final value is subtracted from the initial one ($\Delta w = w_{final} - w_{init}$) such that net weight increase is shown in red while net weight decrease is shown in blue. Each data point in the heat-map corresponds to an average $\Delta w_{hid}$ computed over 20 simulations of the same mean rate pairing (each with different random seeds). In order to limit the outcomes to the specific mean rate pairings being used we ``disable'' the bistability by setting both $\alpha$ and $\beta$ to zero, while $w_{eff}$ is set to \SI{0}{\milli\volt}. As it can be observed, depending on the correlations added to the spike trains in terms of time shift the transition boundary from depression ($\Delta w < 0$) to potentiation ($\Delta w > 0$) can move up (negative shifts), increasing the overall depression region, or down (positive shifts), increasing the overall potentiation region.

\section{Spiking Feed-forward Network}

Being able to recognize regularities in data is one of the core aspects of systems able to learn and adapt, being those biological or artificial. In this section we will describe a basic \gls{sfnn} architecture where the synapses between the units are learned with the \gls{bcall} learning rule.

The pattern recognition task to be carried out here consists of classifying the hand-written digits of the MNIST dataset~\cite{LeCun_etal}. Otherwise simple when compared to other image classification datasets, the MNIST is a commonly used benchmarking tool in the Machine Learning field. The dataset consists of a total of 70000, 28$\times$28 pixels gray-scale images separated between 10 classes (digits 0 to 9), 60000 being used for training and 10000 being used for testing. 

There are several reasons why the MNIST dataset has become a popular benchmark for evaluating new algorithms: well-curated and widely available, making it easy to obtain and use for research purposes; relatively small, which makes it ideal for testing and evaluating new algorithms quickly; challenging enough to test the generalization capability of the algorithms, yet not so complex as to make it computationally intractable. Additionally the dataset has been used extensively in the literature. For these reasons MNIST is also commonly used to train \glspl{snn}.

\begin{figure}[htbp]
\centerline{\includegraphics[width=12cm]{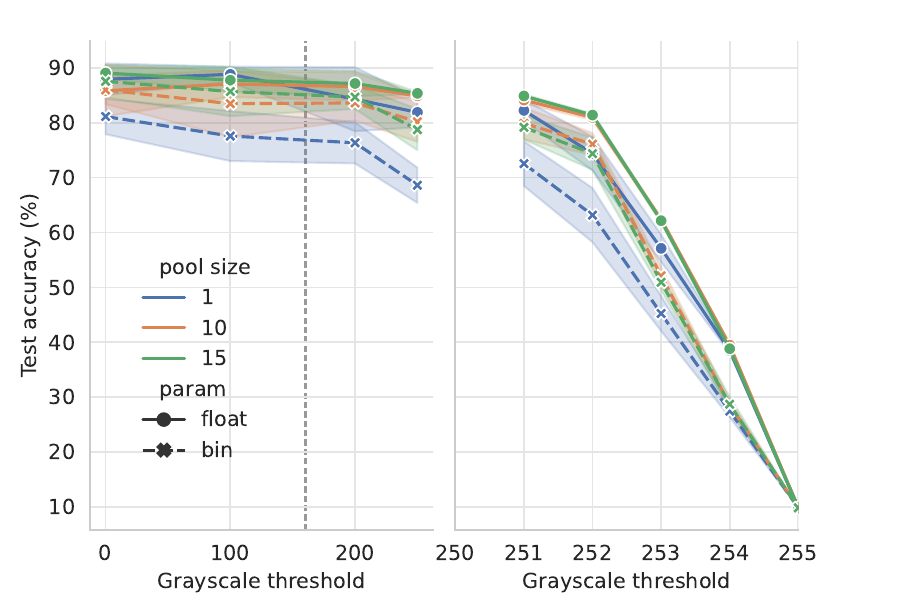}}
\caption{Test accuracy of a Linear Classifier trained on the MNIST dataset as a function of the gray-scale threshold applied to MNIST digits, for different number of units (pool size) allocated to encode each of the 10 output classes, and with floating point (float) or binary network parameters (bin): Mean (lines) and standard deviation (shaded area), over 5 independent parameter initializations. The dashed vertical line indicates the threshold value used in our simulations (i.e., 160).  }
\label{fig:LinearClass_vs_thr}
\end{figure}

In order to simplify further the network to be trained, the gray-scale images in the dataset are binarized by thresholding the pixel intensity to have black and white images. To guarantee that such pre-processing step would not introduce a significant information loss, a linear classifier was trained on MNIST classification, with one-hot encoding of the output class, i.e. with as many output units as the number of output classes, and Cross-Entropy loss. \figurename~\ref{fig:LinearClass_vs_thr} shows the test accuracy of such classifier as a function of the gray-scale threshold. Two factors are varied: the number of units (pool size) allocated to encode each of the 10 output classes, and whether the network parameters are floating points (float) or binarized with quantization-aware training (bin). The dashed vertical line indicates the threshold value used in our simulations (i.e., 160).  In all tested scenarios, no significant performance drop is reported only for threshold values smaller than 250, suggesting that most of the relevant information content for the task at hand is confined in the spatial location of the pixel digits, rather than in their specific gray-scale value. 

\subsection{Network Architecture}

\begin{figure}[htbp]
\centerline{\includegraphics[width=11cm]{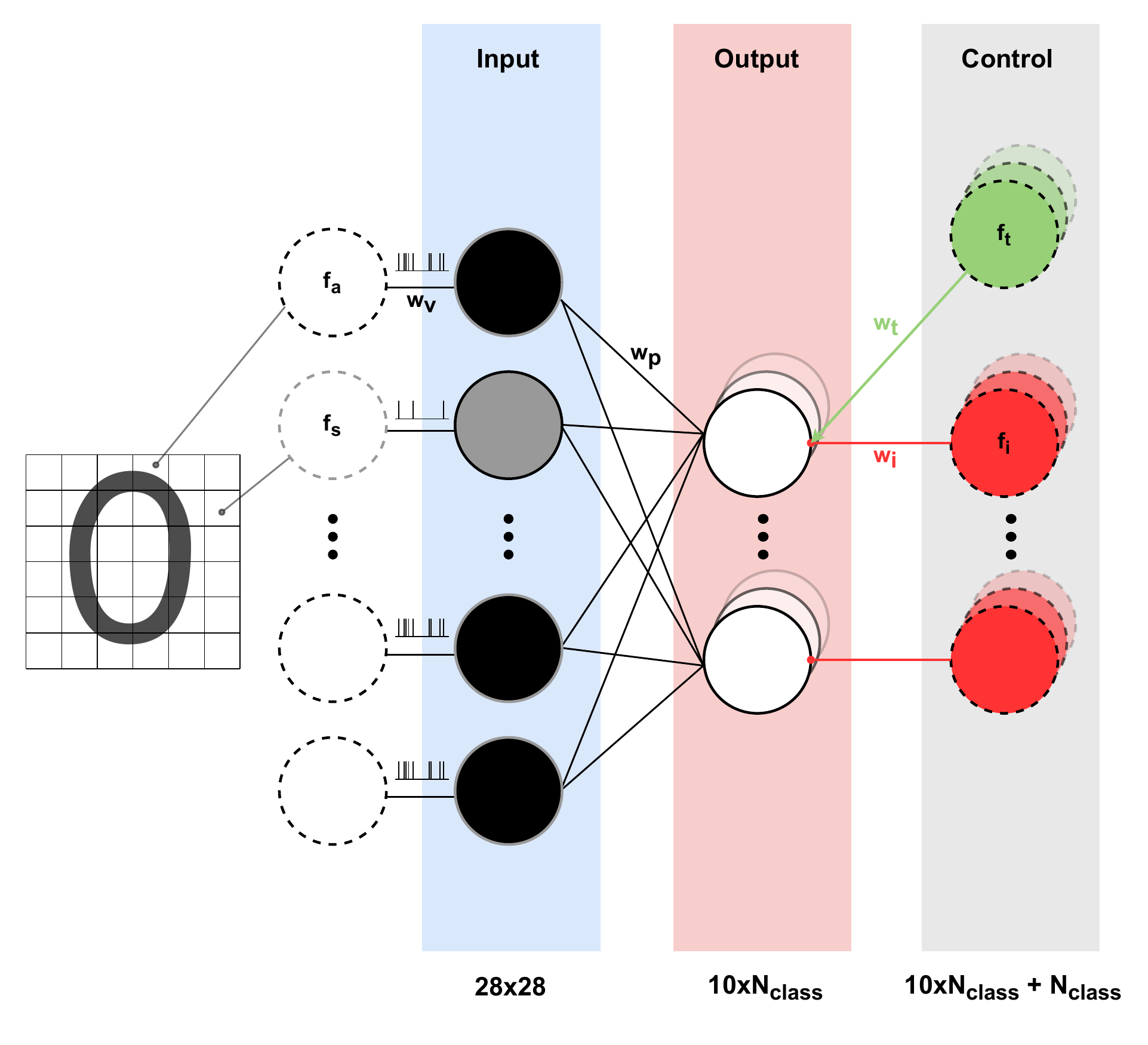}}
\caption{Architecture of the \gls{sfnn} trained for classification with MNIST. Input images are converted into patterns of discrete mean firing rate Poisson spike trains via virtual input units (dashed white circles), making input layer neurons emit spike trains of high (black) or low (gray) mean rate. While virtual units acting as teachers (dashed green circles) provide extra excitatory input to different subsets of output layer neurons depending on the input class, the inhibitory virtual units (dashed red circles) provide constant inhibition to the entire output layer.}
\label{sfnn_topology}
\end{figure}

We implement a simple single layer \gls{sfnn} architecture, that is, a perceptron network: each output unit is computing the weighted sum of its inputs, that is, a direct encoding of the input space. The architecture of the network is depicted in \figurename~\ref{sfnn_topology}.

Since we are working with spiking neuron models (i.e. with stateful units), static inputs (i.e. without a temporal dimension) have to be converted into a spike-train representation. The gray scale MNIST digits are binarized such that pixels with value $0$ are represented by Poisson spike trains of $f_{s}$\SI{}{\hertz}, while pixels with value $1$ are represented by Poisson spike trains of $f_{a}$\SI{}{\hertz}. The units in \figurename~\ref{sfnn_topology} without an internal state (i.e. having the sole purpose of generating Poisson spike trains) are represented by dashed outlines and will be referred to as \textit{virtual neurons}. The virtual input units transform a static image into spike trains of $t_{inp}$ seconds, which are in turn fed 1-to-1 via fixed weight synapses into the input-layer neurons. The weight $w_{v}$ of these connections is set in a way that each spike in the virtual unit causes the respective input neuron to spike.

This discretization of the input frequencies relates to how the network is initialized: all $w_{hid}$ variables are initialized with uniformly distributed random values between $0$ and $1$ such that, on average, $50\%$ of the weights connecting the input to the output layer are set to $w_{eff}$ ($w_{hid} > \theta_{w}$) and $50\%$ are set to \SI{0}{\milli\volt} ($w_{hid} < \theta_{w}$, with $\theta_{w} = 0.5$). As the training unfolds, synapses linked to pixels representing features (black pixels) are strengthened due to a high pre- and post-synaptic mean rate pairing, while synapses linked to empty regions of the image (white pixels) are weakened by paring a low pre-synaptic mean rate with a high post-synaptic one.

The input neurons connect in an all-to-all fashion to output layer neurons (i.e. a fully connected network). We note that the only learnable weights in the network are the ones connecting these two layers. There are at minimum as many output neurons as classes to be learned. Thus, since the network is trained to perform MNIST classification, the smallest output layer size is $10$, with the first output neuron encoding for digits $0$, the second encoding for digits $1$, and so on. A class can also be encoded by a pool (e.g. more than one) of output neurons, which is equivalent to having an ensemble of such networks for classification.

The control layer of the network is composed of multiple pairs of excitatory and inhibitory virtual neurons, one pair per neuron in the output layer. The excitatory units are referred to as ``teacher neurons'' because they provide the class label during training in the form of an additional source of excitation to output neurons (other than the one from the input layer). 

For each input sample presented during training, this teacher input is fed to the pool of output neurons that should learn to respond to the sample class. Boosting the firing rate of the neurons inside this pool potentiates the plastic synapses between the input units with higher activation and the neurons in the target output pool, thereby promoting learning of the right encoding. 

The inhibitory units complement this process of ``class assignment'' by inhibiting all neurons in the output layer. Thus, during training, only the neurons in the output pool receiving the teacher input will have net excitatory input to maintain a higher mean firing rate relative to the rest of the output layer.

As shown in Subsection~\ref{inhibitory_input}, such inhibitory input will also play a role in allowing the learning of patterns with large differences in coding-level, i.e., for binary vectors, the fraction of positions with value one. Thus, as we will show, the inhibitory input is provided during both training and inference.

The training scheme consists of presenting input samples randomly in a sequential manner, for $t_{inp}$ seconds each, and with one presentation for each sample. The hyperparamenters for the \gls{sfnn} shown in \figurename~\ref{sfnn_topology} are listed in Table~\ref{tab:sfnn_variables}.

\begin{table}
    \centering
    \begin{tabular}{c p{0.45\linewidth} c c}
 Hyperparameter & Description & Unit & Value\\
 \midrule
         $f_{a}$ & Mean firing rate representation of pixels with value 1. & \SI{}{\hertz} & 20\\
         $f_{s}$ & Mean firing rate representation of pixels with value 0. & \SI{}{\hertz} & 3\\
         $f_{t}$ & Mean firing rate of virtual teacher neurons. & \SI{}{\hertz} & 30\\
         $f_{i}$ & Mean firing rate of virtual inhibitory neurons. & \SI{}{\hertz} & 210\\
         $w_{v}$ & Weight between virtual input neurons and input layer neurons. & \SI{}{\milli\volt} & 100\\
         $w_{p}$ & Weight between input and output layer neurons. & \SI{}{\milli\volt} & $w_{max}$\\
         $w_{t}$ & Weight between virtual teacher neurons and output layer neurons. & \SI{}{\milli\volt} & 50\\
         $w_{i}$ & Weight between virtual inhibitory neurons and output layer neurons. & \SI{}{\milli\volt} & 30\\
         $t_{inp}$ & Duration of a single sample being inputted to the \gls{sfnn}. & \SI{}{\second} & 1\\
 \midrule
    \end{tabular}
\caption{Hyperparameters of \gls{sfnn}.}
\label{tab:sfnn_variables}
\end{table}

\subsection{Inhibitory Input}\label{inhibitory_input}

Since, due to the nature of the learning rule, weight modification requires activity from both sides of the synapse to occur in either direction, the effect of inhibition is meant to strongly hyperpolarize the neuron but not completely silence it - that is, weight changes can happen even in the absence of the teaching signal. 

Due to the probabilistic nature of the inhibitory spike trains, strong enough input (e.g. at network initialization when many synapses are potentiated) can make an output unit spike, with this probability being inversely proportional to the rate of inhibition. This means that, compared to the activity of an output neuron receiving the teacher signal, the unit receiving only inhibition will fire at a low rate, which is likely to depress synapses that cause it to spike: high pre-synaptic activity (driven by input pattern) paired with low post-synaptic activity (lack of extra excitatory input from teacher) leads to an average weight decrease.

\begin{figure}[htbp]
\centerline{\includegraphics[width=12cm]{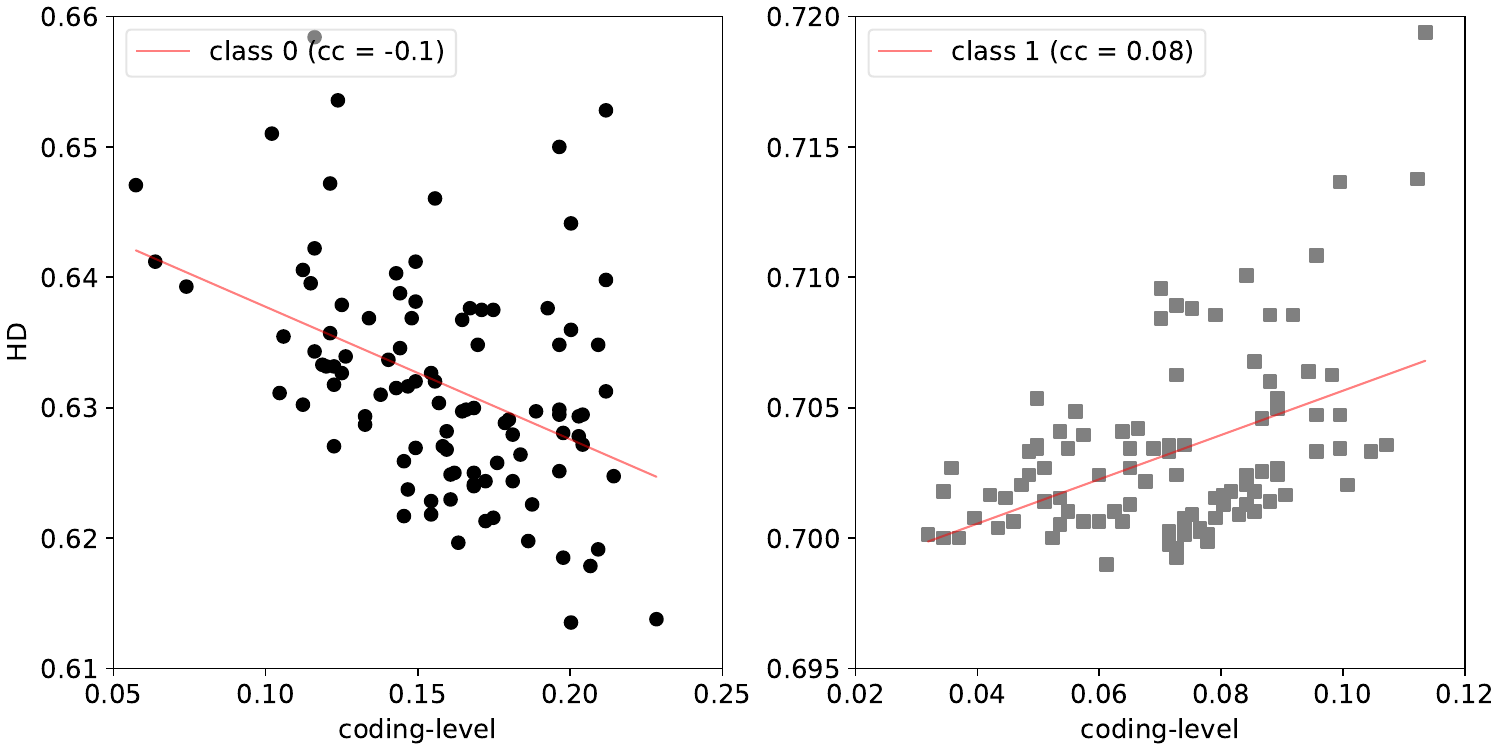}}
\caption{Average Hamming distances between training samples and their assigned output neuron's weight matrices: samples of class $0$ are compared to the binary weight matrix of output neuron index $0$ (left) and samples of class $1$ are compared to the binary weight matrix of output neuron index $1$ (right). Training with fixed inhibition. Distances averaged over 10 independent simulations.}
\label{hammingdist_constantinhibit}
\end{figure}

A naive way of providing this inhibition is by fixing\footnote{For simulations with fixed mean rate inhibition the $f_{i}$ value in Table~\ref{tab:sfnn_variables} is scaled by $0.19$.} the mean firing rate of the inhibitory neurons connected to the output of the network. These neurons are always active, with the excitatory teacher neuron connecting only to output neuron pool being reinforced to respond to a specific class. We show the effect of this setup in training a network with only two classes in \figurename~\ref{hammingdist_constantinhibit}, where the average \gls{hd} between the learned synaptic matrices (the binary weights for the respective class) to each data point used during training is calculated. Simulations in this section are carried out using the network hyperparameters listed in Table~\ref{tab:sfnn_variables} and \gls{bcall} hyperparameters listed in Table~\ref{tab:hyperparams_table}, although we note that for the this subsection the stop-learning mechanism is disabled. All following simulations were done using the Python and the Brian2~\cite{Stimberg2019} package.

Each data point in \figurename~\ref{hammingdist_constantinhibit} represents an average \gls{hd} between a single training sample ($N_{tr} = 200$ training samples, with $100$ samples per class) and ten \glspl{sfnn} independently trained with the same subset of the MNIST dataset. This metric shows how well represented (on average) by the learned weights a sample is. When comparing two binary vectors, the \gls{hd} accounts for the number of positions in which the two vectors differ in value - the more similar they are, smaller the distance. The coding-level of each of the $N_{tr}$ training samples is shown in the x-axis. 

\begin{figure}[htbp]
\centering
{{\includegraphics[width=5cm]{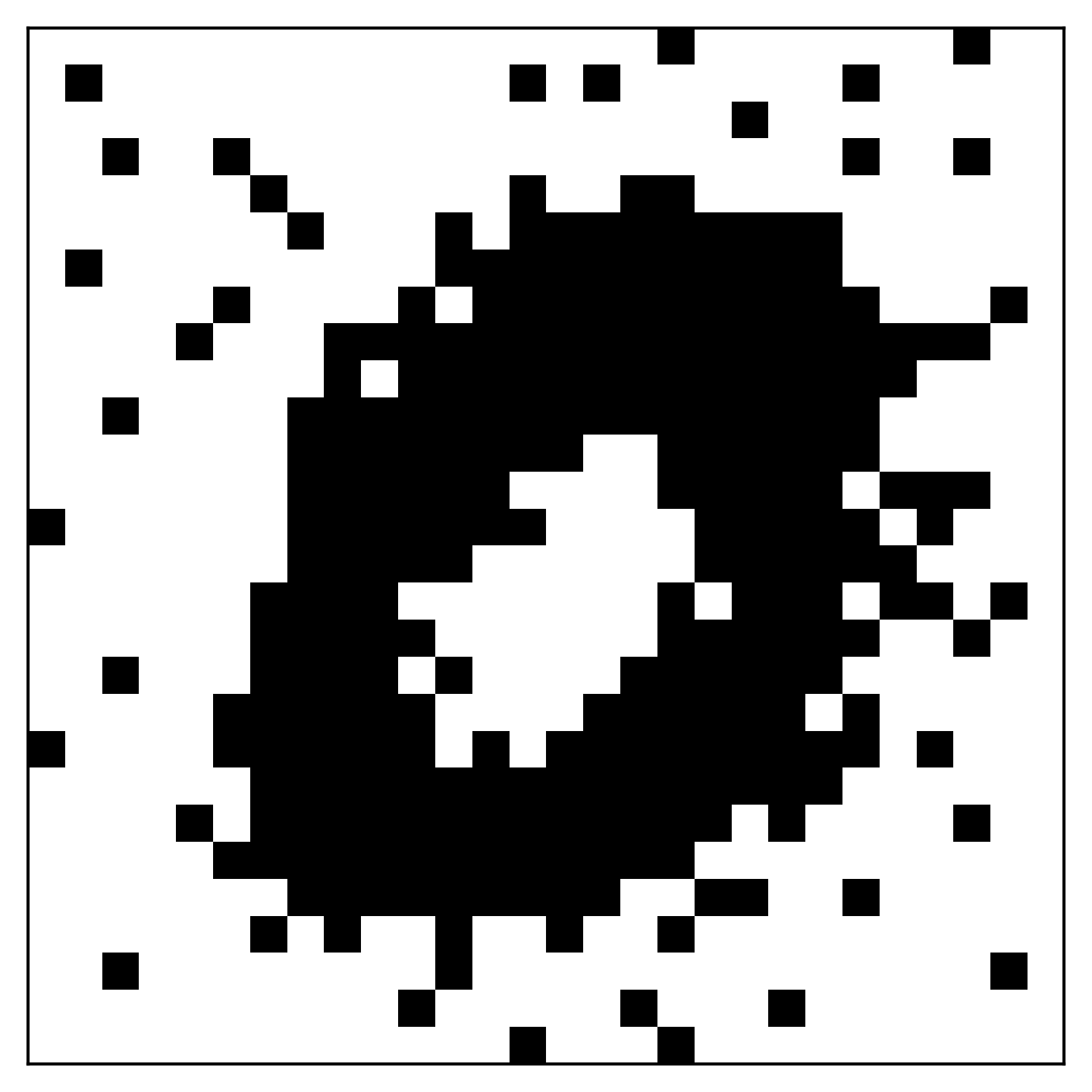}}}
{{\includegraphics[width=5cm]{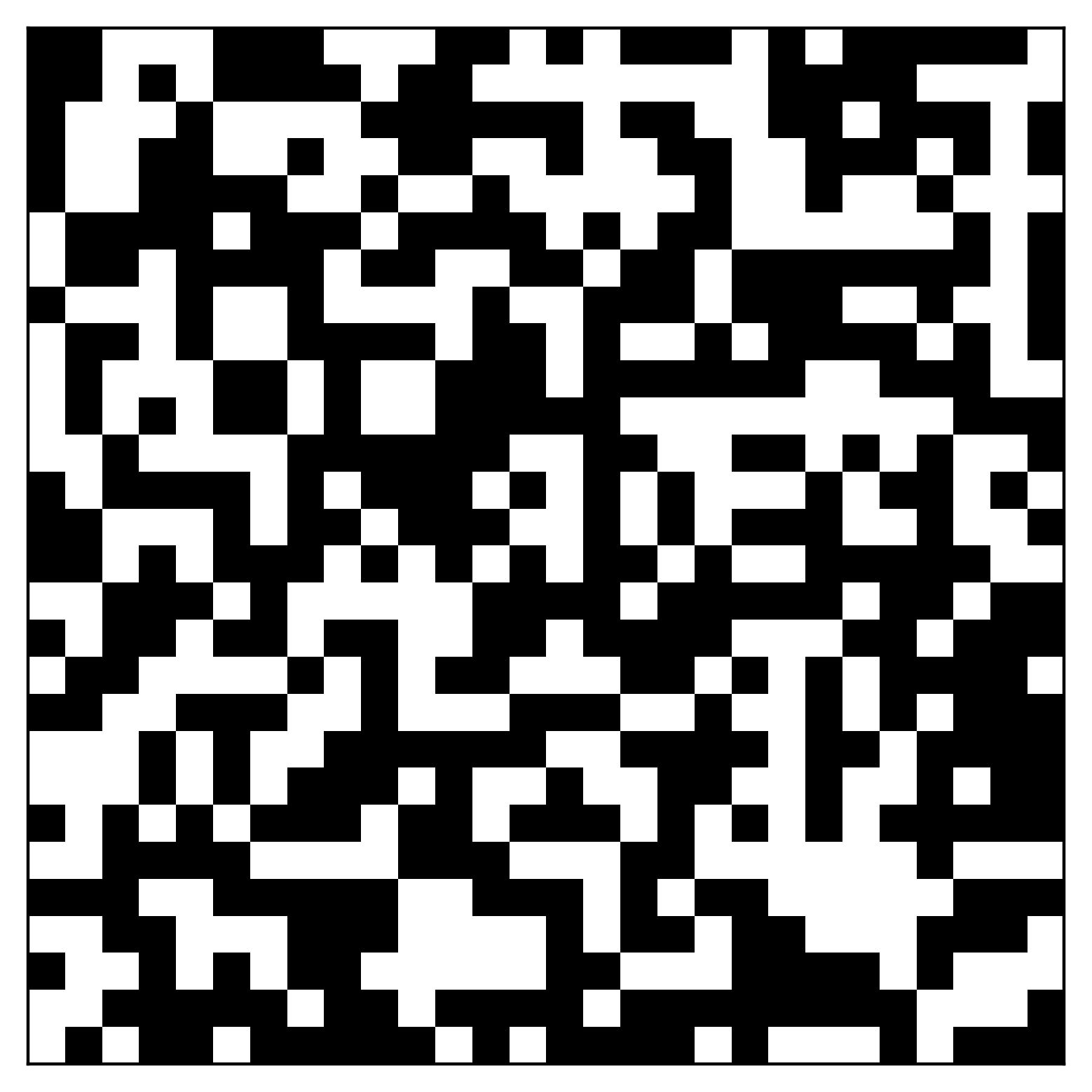}}}
\caption{Learned synaptic matrix for class 0 (left) and 1 (right). Given the differences in coding-level between the two classes an excitation/inhibition balance at the output layer can not be achieved with fixed inhibition.}
\label{learnedSynMatrixConstInhib}
\end{figure}

From the plots in \figurename~\ref{hammingdist_constantinhibit} the first thing to notice is that after training there is a linear relation between how well a sample matches the learned weights and its coding-level. The second observation is that samples of class $0$ seem to better match the learned weight matrices for their class than class $1$ samples do. Two examples of such learned synaptic matrices are shown \figurename~\ref{learnedSynMatrixConstInhib}.

From the differences in range of \gls{hd} values shown in the y-axis of \figurename~\ref{hammingdist_constantinhibit} and the sampled weight matrices shown in \figurename~\ref{learnedSynMatrixConstInhib} one can see that samples of class $0$, having higher coding-level when compared to patterns of class $1$ (relative percentage difference of $\approx 78\%$), are properly represented (left plot). On the other hand, the learned synaptic matrices corresponding to class $1$ (right plot) look random. This means that the total synaptic input due to a presentation of a sample from this class plus the teachers input is not enough to drive the respective output neuron into a mean firing rate high enough to imprint the features of the samples of this class into its synaptic matrix.

If the amount of inhibition provided to the output layer were to be adjusted down to allow class $1$ to be learned, the mean firing rate of the output neuron encoding class $0$ would increase, which would lead it to retain much of the randomly initialized weights that do not correlate with features from class $0$. In the specific case we show in \figurename~\ref{hammingdist_constantinhibit} and \figurename~\ref{learnedSynMatrixConstInhib}, whatever balance is tuned between teacher and inhibitory input for a synaptic matrix with coding-level around 0.16 (e.g. average coding-level of digits 0) will not be the same once the excitation from the input layer drops to half of it (e.g. average coding-level of digits 1). 

This is one of the problems arising from training a single layer \gls{sfnn} with binary synapses to classify patterns with considerable inter-class coding-level differences: the output layer needs to dynamically adjust the excitation/inhibition ratio in order to make up for differences in total synaptic input that samples from different classes can provide.

In the networks we describe, this problem can be solved by making the inhibition directly modulated by the input's coding-level. In this coding-level-dependent inhibition configuration, the inhibitory mean rate is proportional to the input's coding-level\footnote{For simulations with coding-level-dependent inhibition the $f_{i}$ value in Table~\ref{tab:sfnn_variables} is scaled by the coding-level of the sample being presented.}. This would allow for samples with high coding-level that are able to provide ``strong'' excitation will increase the inhibition seeing by the output unit learning to encode that class, while samples with small coding-level will lower the inhibition enough to compensate for the weaker excitatory input from the input layer.

\begin{figure}[htbp]
\centerline{\includegraphics[width=12cm]{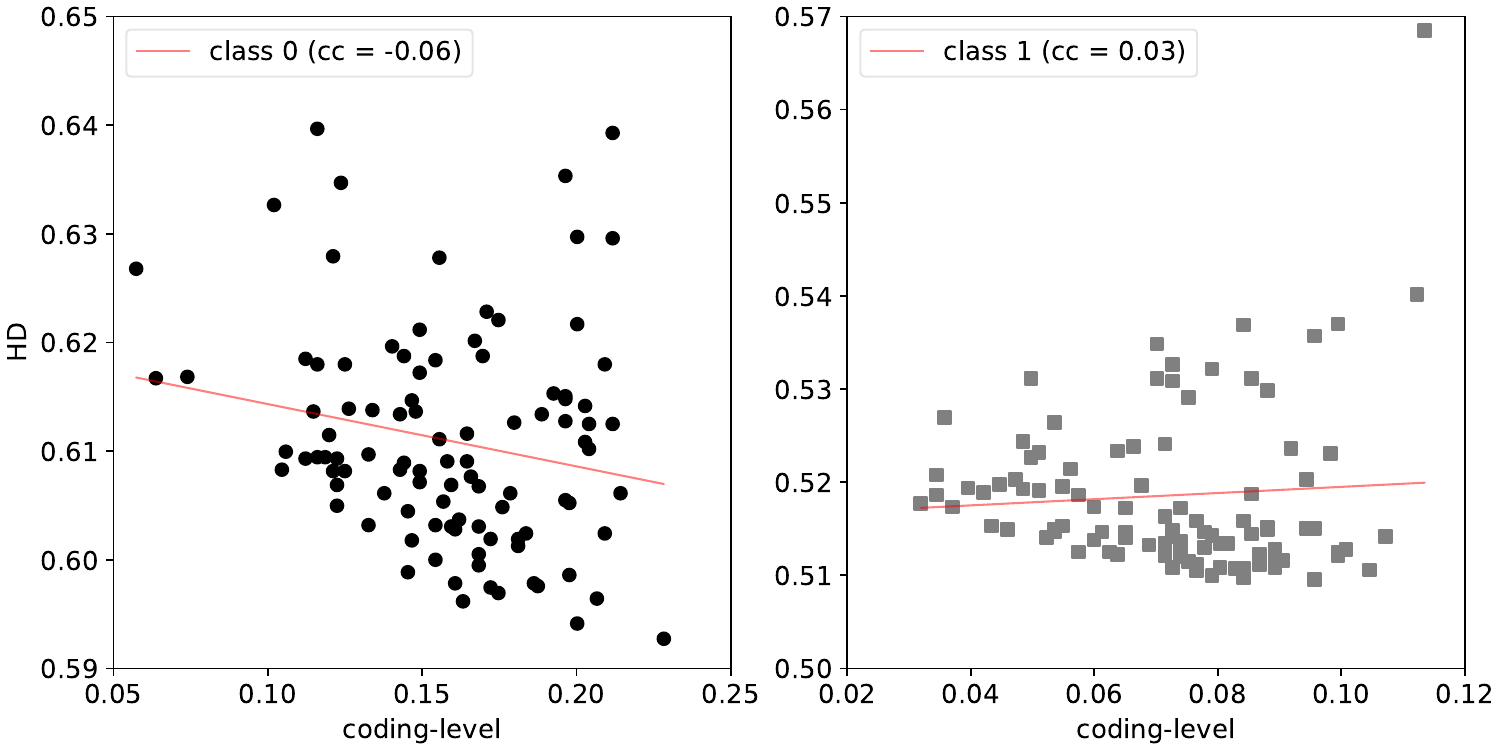}}
\caption{Average hamming distances of training data points to the respective learned prototypes over data point coding level for class of digits 0 (left) and 1 (right). Training with coding level-dependent inhibition. Distances averaged over 10 independent simulations.}
\label{hammingdist_dyninhibit}
\end{figure}

\begin{figure}[htbp]
\centering
{{\includegraphics[width=5cm]{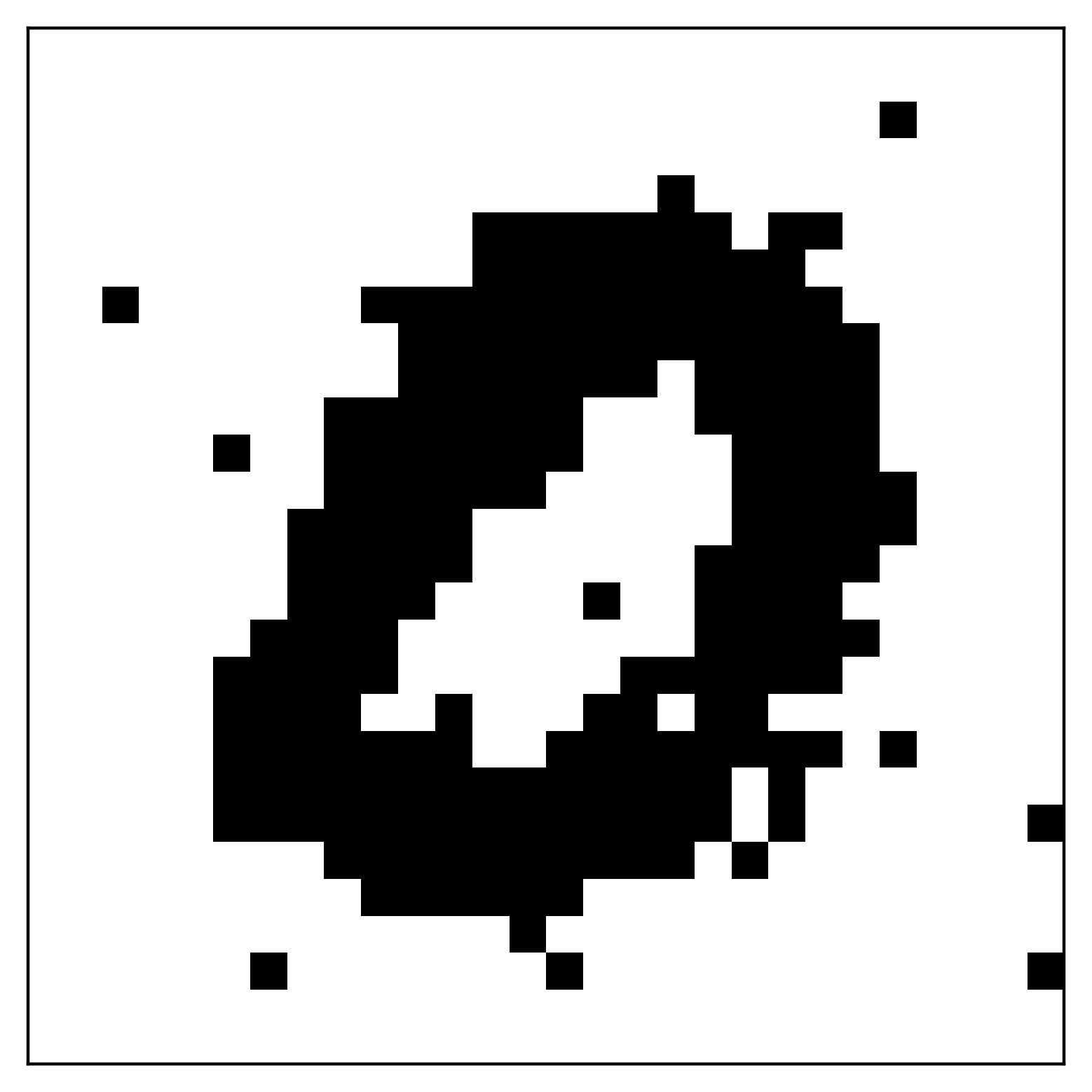}}}
{{\includegraphics[width=5cm]{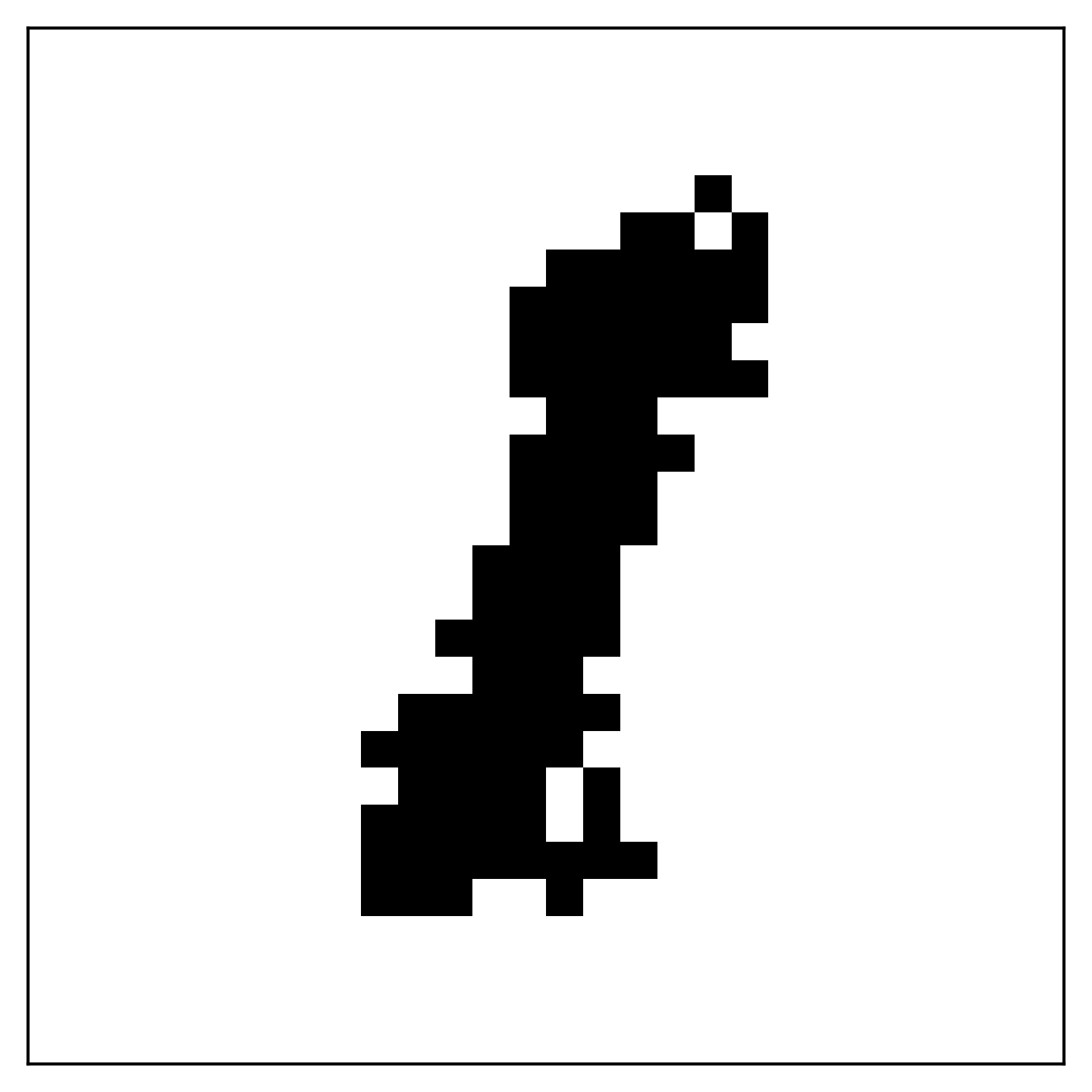}}}
\caption{Learned synaptic matrix for class 0 (left) and 1 (right). Training with coding-level-dependent inhibition allows the excitation/inhibition balance to be dynamically adjusted based on the sample's coding-level, which enables the network to properly learn both classes given the same number of samples.}
\label{learnedSynMatrixDynInhib}
\end{figure}

The results of training our networks using this coding-level-dependent inhibition scheme are shown in \figurename~\ref{hammingdist_dyninhibit} and \ref{learnedSynMatrixDynInhib}. As expected, we see that class 1 is now represented in the learned synaptic matrix similarly to what happens to class 0.

Our simulations show that this modulated inhibition is necessary to properly encode (train) patterns with varying coding-level. We look into whether this coding-level-dependent inhibition is relevant beyond training by testing \glspl{sfnn} with and without it during inference. Similarly to the previous simulations, each data point measures the average mean rate response to the same $N_{ts} = 20$ ($10$ samples per class) testing samples for 10 independently trained and tested \glspl{sfnn} on both settings.

\begin{figure*}[t]
 \centerline{\includegraphics[width=\linewidth]{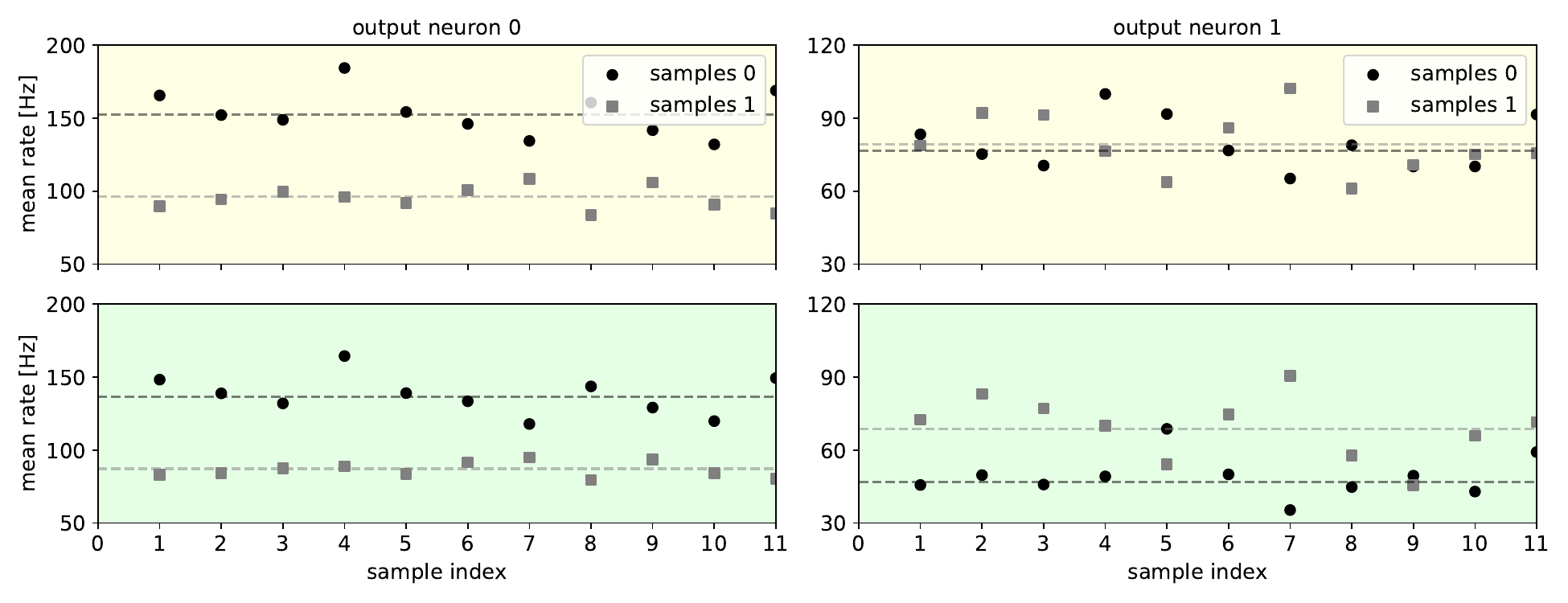}}
\caption{Mean rate responses of each output neuron to ten samples of each class (0s and 1s). The plots contrast the separability of the neurons' responses with (green background) and without (yellow background) coding-level-dependent inhibition during inference. Responses are averaged over 10 independent simulations for each setting.}
\label{testedDynMatrixDynInhib}
\end{figure*}

In \figurename~\ref{testedDynMatrixDynInhib} we show how the mean rate responses each output neuron gives for each class with (green background) and without\footnote{Inhibitory neurons are silenced during inference.} (yellow background) coding-level-dependent inhibition during inference. From it we see that if no inhibitory input is provided during inference (yellow background) the mean rate responses of the output neuron ``responsible'' for the class with low coding-level are similar for both classes. This analysis show that this dynamic modulation of inhibition at the output layer for the networks described here is not only necessary to properly learn a class representation but also to increase the selectivity in the responses learned to patterns with low coding-level.

\subsection{Constrained Maximum Response}

Learning a class prototype, that is, a representation for a specific group of patterns, relies on potentiating synapses between one output layer neuron $j$ and input neurons that are active for that class throughout training, such that frequent features of the patterns shown are imprinted on the synaptic matrix $W_{j}$ connecting the input layer to output neuron $j$. 

As it can be seen in \figurename~\ref{learnedSynMatrixDynInhib}, the number of non-zero elements in $W_{j}$ depends on the coding-level of patterns in a class. This means that, with the constraint of binary synapses, the maximum firing rate of an output neuron is constrained by the coding-level of the patterns it learns to be responsive to. This is the second problem arising from differences in coding-level. 

The effect of this constrained response can be see in \figurename~\ref{testedDynMatrixDynInhib}: independently of providing or not inhibitory input during inference, we see by comparing the plots on the left to the plots on the right that the mean rate responses of the output neuron encoding class 0 (approximately between \SI{100}{\hertz}-\SI{150}{\hertz}) are higher than the responses of the output neuron encoding class 1 (approximately between \SI{60}{\hertz}-\SI{90}{\hertz}). Even though the weight matrices correctly represent the digits (see \figurename~\ref{learnedSynMatrixDynInhib}) it is clear that the limits on the mean rate responses imposed by the differences in the coding-level of the classes will prevent the proper classification of new samples: even though each neuron learns separable responses for each class, output neuron 0 is able to respond to samples of class 1 with a mean firing rate that is about \SI{20}{\hertz} higher than what output neuron 1 is able to give for the same samples.

Similar to the approach adopted in~\cite{Brader_etal2007} we use a stop-learning mechanism to make the network's response invariant to the coding-level of patterns being learned. The upper threshold $\theta_{h}$ of the stop-learning imposes a limit to the mean rate for which weight change is possible. Once the total synaptic input is effective in driving the neuron to this mean rate threshold, the instantaneous weight configuration for that neuron is stabilized by the bistability while $x_{s}$ is outside thresholds $\theta_{h}$ and $\theta_{l}$. In what follows we show how this is relevant for reducing this gap in response shown in \figurename~\ref{testedDynMatrixDynInhib}.

With randomly initialized weights, during the initial stages of training a neuron's response is ``weak'' since, probabilistically, only about $50\%$ of the synapses are capturing features of the input pattern. As training progresses, the weights start to capture the pattern of activity on the input layer and the neuron begins to respond with higher mean firing rates as more synapses are recruited (i.e., cross $\theta_{w}$). As the weight matrix of an output neuron begins to represent the class it is learning, the excitatory input experienced by the neuron increases to the point where its stop-learning trace $x_{s}$ starts to go beyond $\theta^{h}$ whenever a sample of that class is shown. That is, while a neurons $x_{s}$ trace oscillates well between $\theta^{l}$ and $\theta^{h}$ with randomly initialized weights, with a synaptic matrix that captures the input activity for a specific class this trace will overshoot beyond $\theta^{h}$ whenever a sample of such class is shown, which slows down learning and stabilizes the modifications made to the synaptic matrix.

\begin{figure}
    \centering
    {{\includegraphics[width=5cm]{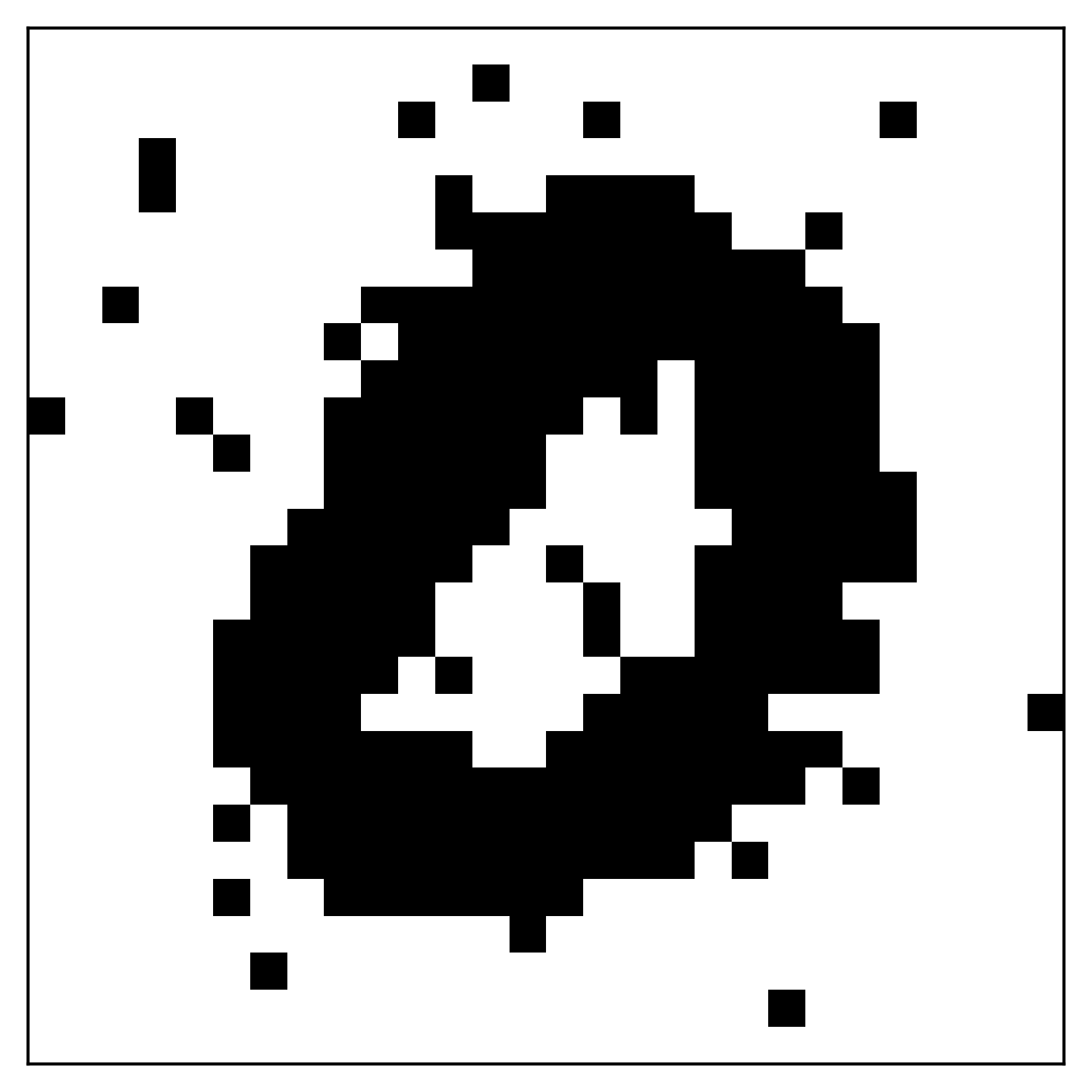}}}
    {{\includegraphics[width=5cm]{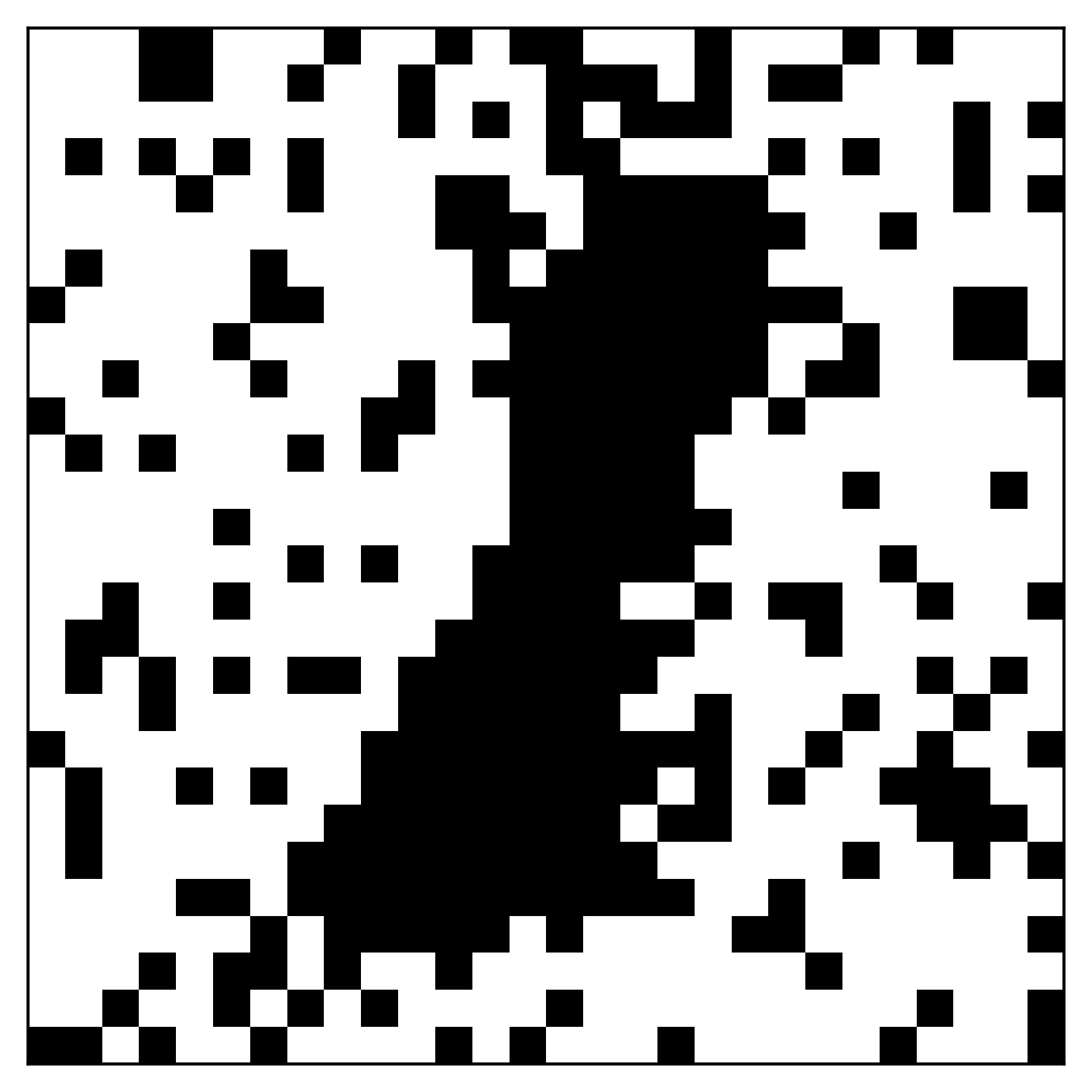}}}
    \caption{Learned synaptic matrix for class 0 (left) and 1 (right). Training with both coding-level-dependent inhibition and the stop-learning mechanism allows the output neuron encoding for the class with low coding-level to learn the features from the samples while retaining some of the randomly initialized weights that help in increasing its baseline mean firing rate (i.e. the mean firing rage caused by the features of the samples alone).}
    \label{stopTranceTrainingWMatrices}
\end{figure}

Pairing this stop-learning trace dynamics with coding-level-dependent inhibition leads to a faster ``exit'' of the learning window (i.e., $x_{s}$ crossing $\theta^{h}$ from bellow) during training for the neurons encoding classes with relatively low coding-level, which tends to retain randomly-initialized weights that would otherwise be ``erased'' during learning (since these do not correlate with class features). This effect can be seen in the synaptic matrix for class 1 in \figurename~\ref{stopTranceTrainingWMatrices}.

\begin{figure*}[t]
    \centering
    \centerline{\includegraphics[width=\linewidth]{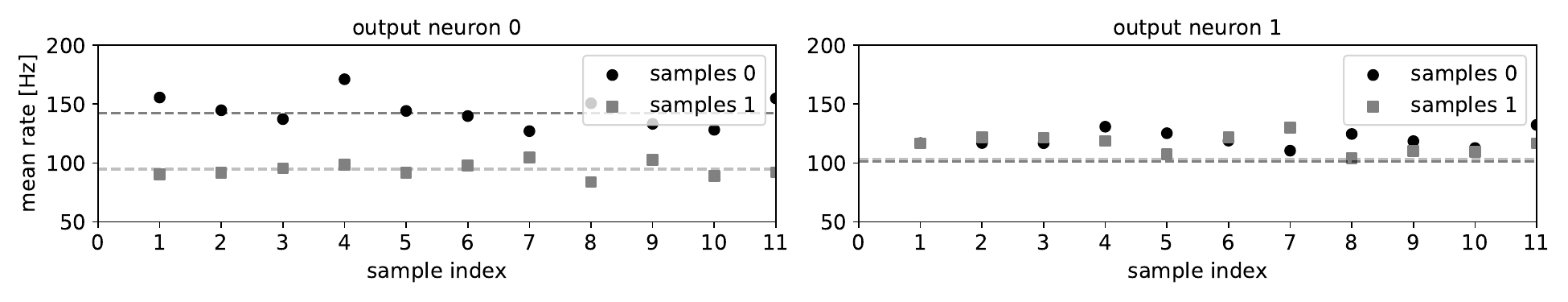}}
    \caption{Mean rate responses of each output neuron to ten samples of each class (0s and 1s). Neurons are trained with the stop-learning mechanism and coding-level-dependent inhibition (during training and inference). Responses are averaged over 10 independent simulations for each setting.}
    \label{codinglevelplusstoplearning}
\end{figure*}

Comparing \figurename~\ref{testedDynMatrixDynInhib} (training with coding-level-dependent inhibition) with \figurename~\ref{codinglevelplusstoplearning} (training with coding-level-dependent inhibition and the stop-learning mechanism) we see that now that, although there are still small differences, the mean rate responses for both neurons are within similar ranges. More importantly, between these two neurons, output neuron 0 has the highest average mean firing rate to samples from class 0 and output neuron 1 has the highest average mean firing rate to samples from class 1. 

We note one unwanted side-effect of this solution to the coding-level problem: while the ``retained noise'' (from the random initialization of the weights) in the learned matrix for class 1 increases its neuron`s mean firing rate it also increases correlations with other class by increasing potential overlaps between them (e.g. some of these ``noise'' weights might match features from other classes). This can be seen in how the average mean rate responses for output neuron 1 shown in \figurename~\ref{codinglevelplusstoplearning} are much closer between the two classes when compared to how separable they are when the stop-learning mechanism is not utilized (see \figurename~\ref{testedDynMatrixDynInhib}).

\subsection{Dynamics Transients}

In real-world scenarios the time interval between two data points can depend on several factors (e.g. how data is being collected by a sensor, buffering, etc). This raises the question of how to input data into SNNs. 

Due to the time dynamics involved in a spiking network computation, a neuron response at a time step ($t^{i}$) will depend its response at previous time steps ($t^{i-1},t^{i-2},...$), the same can be said about the traces used in the learning rule. This means that, upon feeding some data into the network, transient activity due to the recent past of activation of the network's units will interfere with the elicited activity to a new input. 

Here we look into how training is affected by inputting patterns into the network in two different ways. In the first, a period of silence (i.e. only fixed inhibitory input being active) of 1 second where no input is provided to the input layer is enforced between any two patterns. In the second, the patterns are shown continuously without delays in between. We look into how this difference affects learning by tracking the weight evolution of synapses encoding pixels that overlap between the classes of digits 0 and 1. Given that potentiating synapses representing correlated pixels of different classes would decrease classification performance by making the learned representations less orthogonal, minimizing the number of such synapses learned would be beneficial for better classification accuracy. 

\begin{figure}[htbp]
\centerline{\includegraphics[width=10cm]{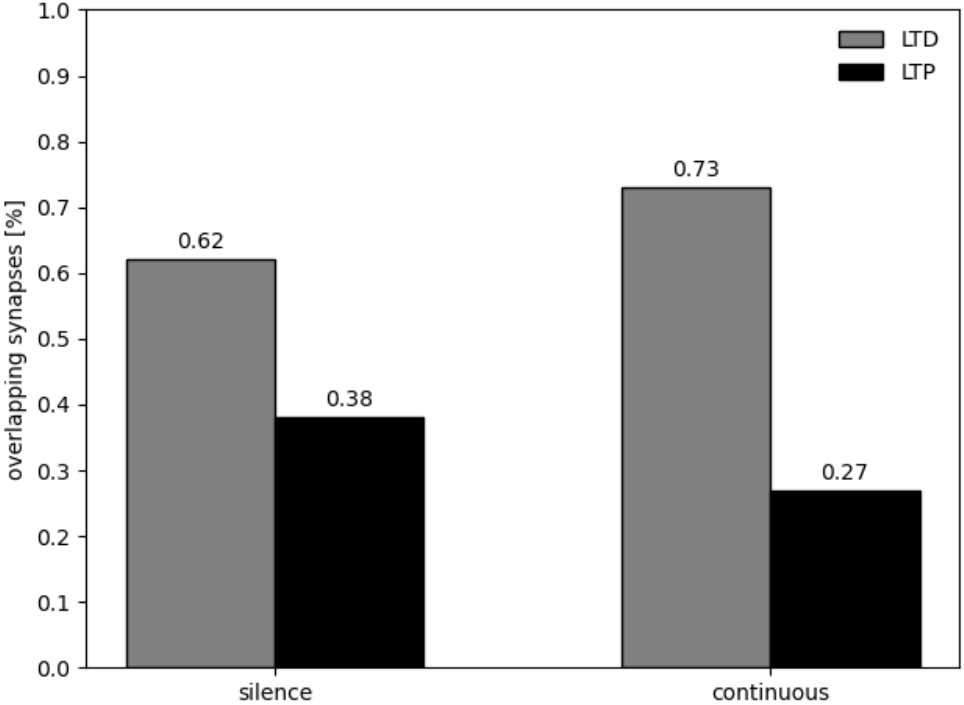}}
\caption{Comparison between training with periods of no activity in the network between different class presentations (left) and continuous presentations (right). The average percentage of potentiated synapse representing overlapping pixels between patterns of two different classes is calculated over 20 independent network simulations with reduced number of training and testing data points.}
\label{overlappingSynapses}
\end{figure}

In \figurename~\ref{overlappingSynapses} we show the proportions of potentiated and depressed synapses after learning with 80 data points (40 per class). The overlapping percentage is estimated by tracking the same overlapping pixels over 20 independent simulated networks and averaging the percentage of synapses related to such pixels that where potentiated/depressed.

We verify that presenting the patterns continuously to the network on average translates into learned synaptic matrices in which the overlaps in the learned representations are minimized by having a higher percentage of depressed overlapping synapses. During training, an output neuron $a$ responding to its class at time step $t^{i}$ will have a high calcium concentration. If at time step $t^{i+1}$ the target output neuron changes, the mean rate of neuron $a$ will be lowered, which will biases synaptic updates towards depression events (high mean rate at the input paired with low mean rate at the output). After switching the input class, if neuron $a$ is still able to emit spikes this means that part of the currently presented input activates synapses to $a$ (i.e. overlapping pixels between samples), so such synapses will be depressed given that $a$ is spiking with a low mean rate. 

\begin{figure}[htbp]
\centerline{\includegraphics[width=9cm]{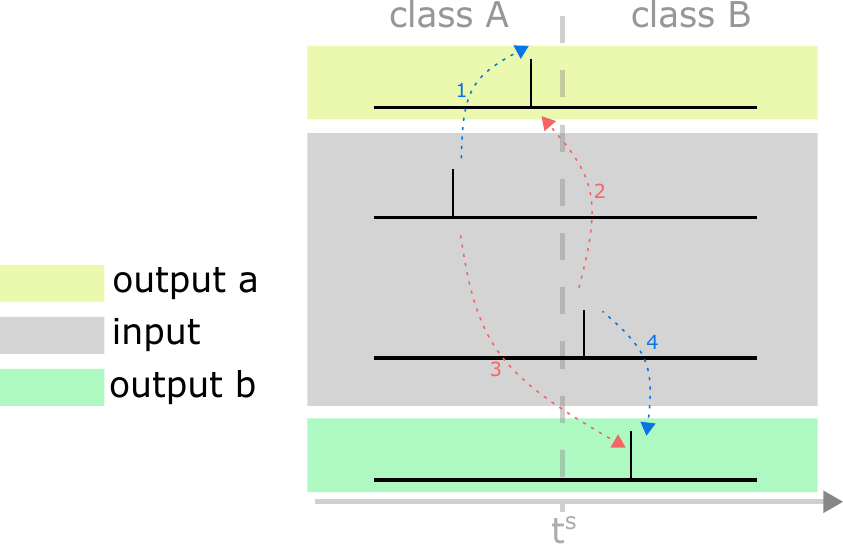}}
\caption{Illustration of transients (traces' change due to spiking) interaction at time $t^{s}$ where the input class is switched. The read out of calcium traces shortly before and after $t^{s}$ decorrelates the synaptic matrices by depressing pre-post interactions that are distal (red 3) or close (red 2) in time. These spike-pair interactions are not possible if the ongoing dynamics due to input is reset between inputs.}
\label{ex_transients_interaction_sfnn}
\end{figure}

We exemplify in \figurename~\ref{ex_transients_interaction_sfnn} how this pairing of samples is further useful for training. At time $t^{s}$ the input to the network switches from a sample of a class A to a sample of a class B. Since a new output unit is now being driven to learn the pattern, upon spiking right after time $t^{s}$ this unit will see low values for the calcium traces of input units active before $t^{s}$ (red arrow 3) so weights will be depressed according to Eq.~\ref{eq:pre_post_pairing_eq}. At the same time, pre-synaptic spikes encoding class B after $t^{s}$ will readout a high concentration of calcium from the active output unit before $t^{s}$ (red arrow 2), which will lead to a decrease of $w_{hid}$ according to Eq.~\ref{eq:post_pre_pairing_eq} - this allows the output unit after $t^{s}$ to learn to decrease its selective to pixels of a different class.

This decrease in overlap between the learned synaptic matrices can thus be assigned to the interaction between transient activity elicited by samples belonging to neuron's class and subsequent samples of other classes. Interestingly, a previous work~\cite{Illing_etal2019} on spiking networks with a single hidden layer trained on MNIST reported that optimal performance could not be reached in the networks trained if a period of inactivity was not introduced between samples. Here we find the opposite: our networks reach sub-optimal performance if there is no interaction between the activity elicited by samples of different classes. We thus train the networks by presenting data points continuously, without periods quiescence in between.

\section{Digits Classification}

We investigate the \gls{cr}, defined as the number of correctly classified samples over the total number of tested samples, as a function of the output pool size (i.e., the number of output units assigned per class). We read out network activity in two ways: (1) \gls{mr}, where the network prediction corresponds to the class assigned to the output pool containing the neuron with highest activity, or (2) \gls{ar}, where the prediction corresponds to the class assigned to the output pool with the highest average mean firing rate. For each $N_{class}$ we train 10 independent \glspl{snn} and the average \gls{cr} is computed along with its standard deviation.

\begin{table}
    \centering
    \begin{tabular}{c c}
 Hyperparameter & Value\\
 \midrule
         $\tau_{i}$ & \SI{30}{\milli\second}\\
         $\tau_{j}$ & \SI{30}{\milli\second}\\
         $\tau_{s}$ & \SI{800}{\milli\second}\\
         $\tau_{w}$ & \SI{40}{\second}\\
         $a_{i}$ & 0.4\\
         $a_{j}$ & 0.5\\
         $a_{s}$ & 0.075\\
         $\theta_{i}$ & 0.05\\
         $\theta_{j}$ & 0.05\\
         $\theta_{u}$ & 0.55\\
         $\theta_{l}$ & 0.05\\
         $\theta_{w}$ & 0.5\\
         $c_{1}^{d}$ & -0.026\\
         $c_{2}^{d}$ & -0.008\\
         $c^{p}$ & 0.18\\
         $w_{eff}$ & \SI{1}{\milli\volt}\\
         $x^{max}_{i}$ & 1\\
         $x^{max}_{j}$ & 1\\
         $x^{max}_{j}$ & 1\\
 \midrule
    \end{tabular}
\caption{Hyperparameters values used for simulations of \gls{bcall}.}
\label{tab:hyperparams_table}
\end{table}

The hyperparameters in Table \ref{tab:hyperparams_table} are used for training with limited fine tuning. While during training the maximum connection weight $w_{max}$ between input and output layer is set to \SI{1}{\milli\volt}, during testing this parameter is set to \SI{10}{\milli\volt}. The reason for this is that, after training, we sweep this hyper-parameter and find that the \gls{cr} increases with increased $w_{max}$, saturating at 10mV. The connection weights between teacher and output layer, inhibitory neurons and output layer, as well as the mean rates for active and spontaneous input units, and the teacher and inhibitory neurons are shown in Table~\ref{tab:sfnn_variables}. All neuron parameters used for all simulations in this work are shown in Table~\ref{tab:neurons_table}.

\begin{table*}
    \centering
    \begin{tabular}{c p{0.45\linewidth} c c}
 Hyper-parameter & Description & Value exc. & Value inh.\\
 \midrule
         $V_{r}$ & Resting potential of the neuron membrane. & \SI{-65}{\milli\volt} & \SI{-60}{\milli\volt}\\
         $V_{rse}$ & Reset potential of the neuron membrane. & \SI{-65}{\milli\volt} & \SI{-60}{\milli\volt}\\
         $V_{thr}$ & Membrane threshold for spike emission. & \SI{-58}{\milli\volt} & \SI{-40}{\milli\volt}\\
         $V_{incr}$ & Amplitude of membrane threshold adaptation. & \SI{5}{\milli\volt} & -\\
         $\tau_{thr}$ & Time constant of the membrane threshold adaptation. & \SI{20}{\milli\second} & -\\
         $\tau_{mem}$ & Time constant of the neuron membrane. & \SI{20}{\milli\second} & \SI{10}{\milli\second}\\
         $\tau_{epsp}$ & Time constant of the \gls{epsp}. & \SI{3.5}{\milli\second} & \SI{3.5}{\milli\second}\\
         $\tau_{ipsp}$ & Time constant of the \gls{ipsp}. & \SI{5.5}{\milli\second} & \SI{5.5}{\milli\second}\\
 \midrule
    \end{tabular}
\caption{Neuron model parameters.}
\label{tab:neurons_table}
\end{table*}

\begin{figure}[htbp]
\centerline{\includegraphics[width=10cm]{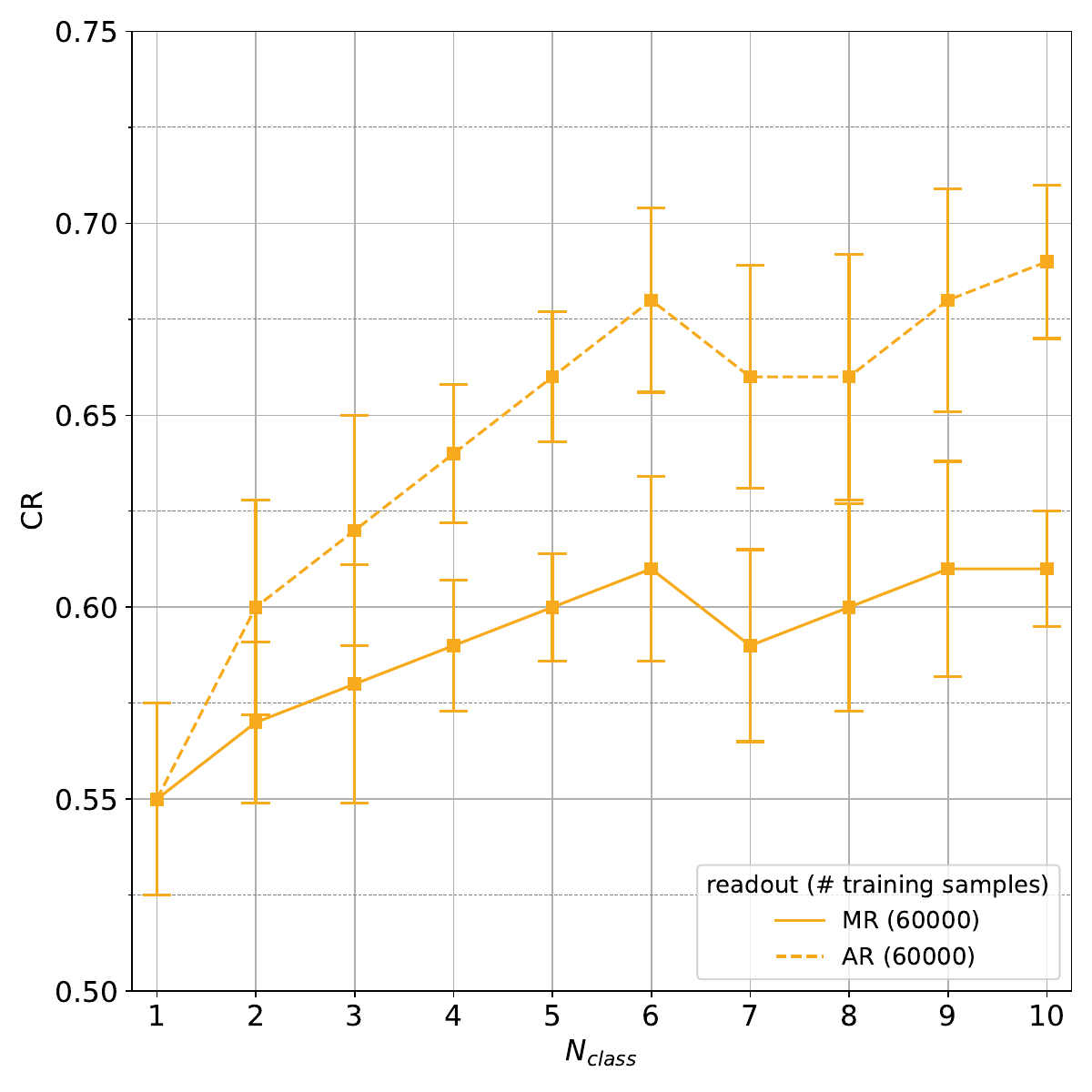}}
\caption{Network performance as a function of output pool size and readout method. Error bars show $\pm1$ standard deviation. \gls{bcall} and \gls{sfnn} hyperparameters in tables \ref{tab:hyperparams_table} and \ref{tab:sfnn_variables}, respectively.}
\label{09_BCaLL_CR_poolsize_scaling}
\end{figure}

\figurename~\ref{09_BCaLL_CR_poolsize_scaling} illustrates the scaling of performance with the output pool size. For the smallest networks with $N_{class} = 1$ (both readout methods are equivalent), the average \gls{cr} is approximately $0.55$. However, the performance of the largest networks varies depending on the readout method. The best result, approximately $0.68$, is achieved when reading the average activity of the readout pool (\gls{ar} method). In contrast, using the maximal mean rate to encode the network prediction (\gls{mr} method) yields a performance of around $0.61$ (an approximate $10\%$ decrease in performance).

\begin{figure}[htbp]
\centerline{\includegraphics[width=10cm]{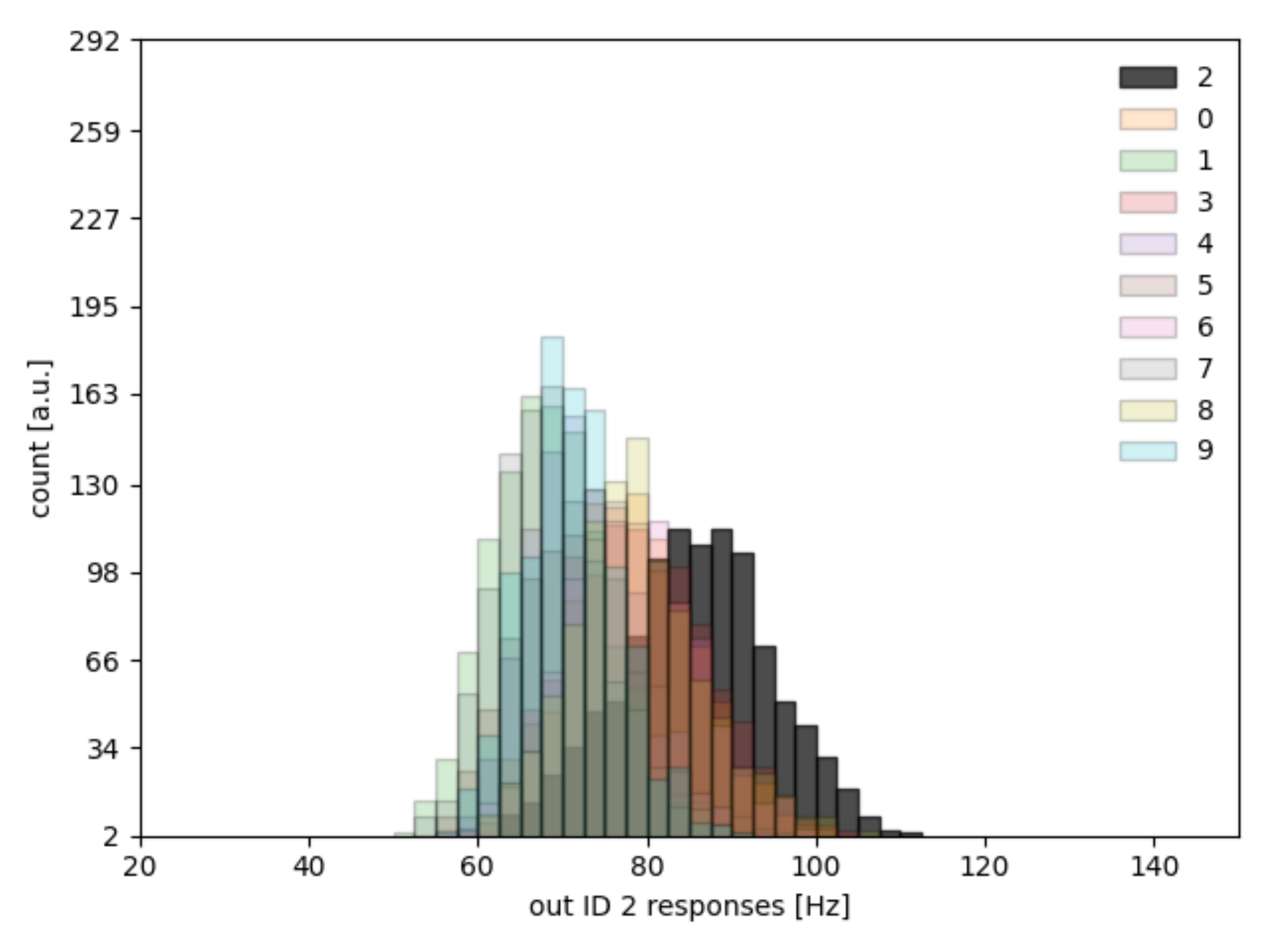}}
\caption{Mean rate response histogram to all classes for an output neuron encoding for class 2 after training. The positive class is shown in gray. The center of the distributions for the negative classes are shifted to the right of the positive class, indicating that the neuron became selective to digits 2.}
\label{mean_rate_response_dist}
\end{figure}

The plot suggests that averaging the readout pool is a more effective approach for extracting the network's prediction when $N_{class} > 1$. To understand why, we examine the selectivity of a sampled output neuron in \figurename~\ref{mean_rate_response_dist}. The histogram displays the mean rate responses of the neuron to the class it learned to encode (shown in gray) compared to its responses to other classes (shown in colors). The neuron exhibits selectivity towards digits 2, as evidenced by the distribution's mean being larger for samples of that class than for other classes. However, due to the non-disjoint nature of these distributions, the neuron may exhibit higher activity in response to some of the non-assigned classes' samples compared to some of its assigned class ones. By averaging the responses within the pool, we reduce the likelihood of misclassifications since a single incorrect high response to a negative class sample (i.e., the \gls{mr} method) within the pool will be mitigated by the overall averaging effect.

Another information given in \figurename~\ref{09_BCaLL_CR_poolsize_scaling} is that performance gains seems to start saturating as $N_{class}$ increases. Increasing the output layer beyond $N_{class} > 1$ in such \glspl{sfnn} is equivalent to using multiple Perceptrons per class, commonly known as an ensemble method in machine learning literature. For this reason, the increase in performance by adding a second output neuron to encode a class can be associated with the fact that this second representation might pick up on features that are different (novelty) from the first output - that is, it can potentially better match variations within the class. This convergence seen in the plot seems to indicate that there is not enough novelty being added to the pool. To check whether or not this is the case, we retrain the networks using 20k training samples out of the original 60k available.

\begin{figure}[htbp]
\centerline{\includegraphics[width=10cm]{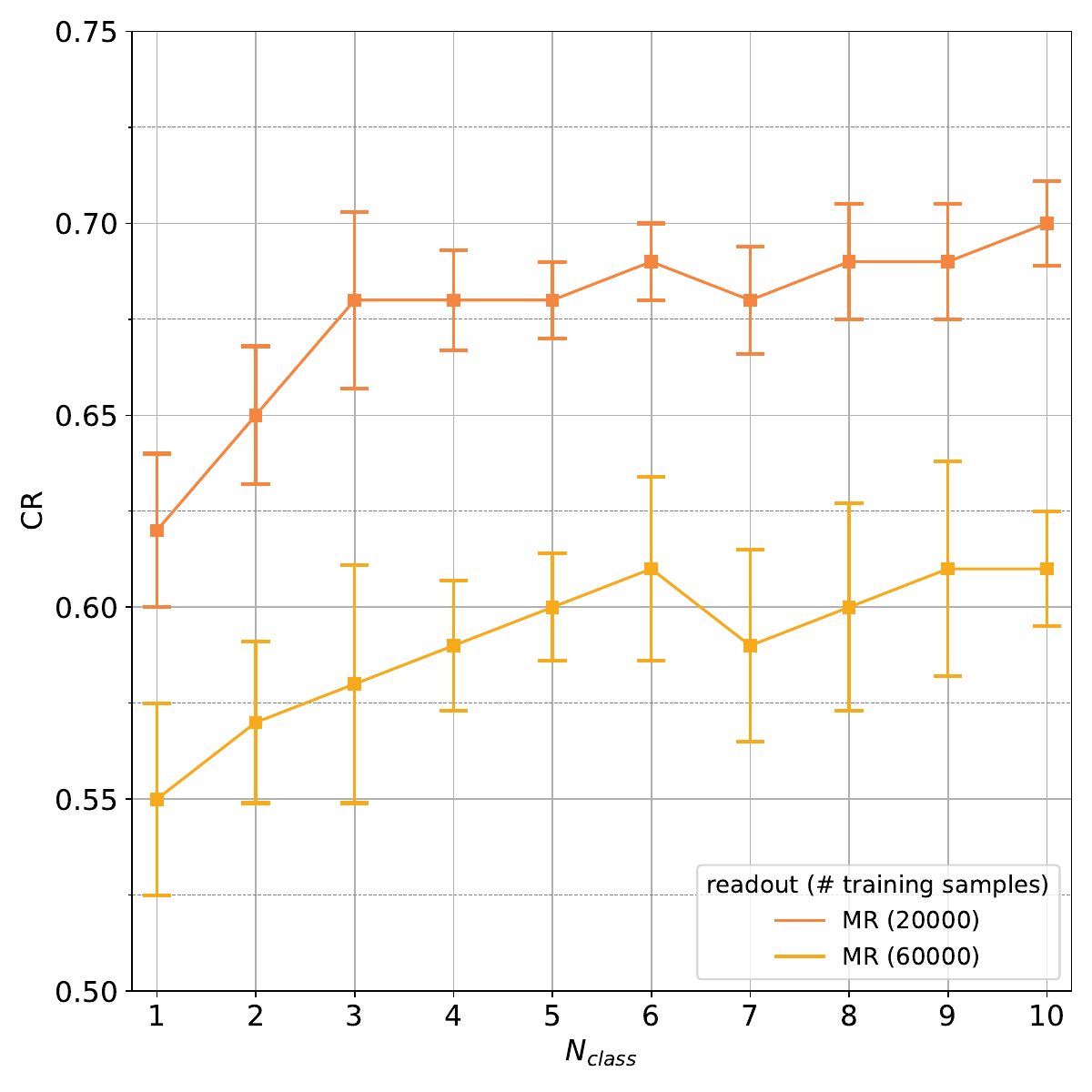}}
\caption{Comparison of network performance as a function of output pool size \gls{mr} voting method between networks training with 60k and 20k training samples. Error bars show $\pm1$ standard deviation. \gls{bcall} and \gls{sfnn} hyperparameters in tables \ref{tab:hyperparams_table} and \ref{tab:sfnn_variables}, respectively.}
\label{poolmax_60k_x_20K}
\end{figure}

\begin{figure}[htbp]
\centerline{\includegraphics[width=10cm]{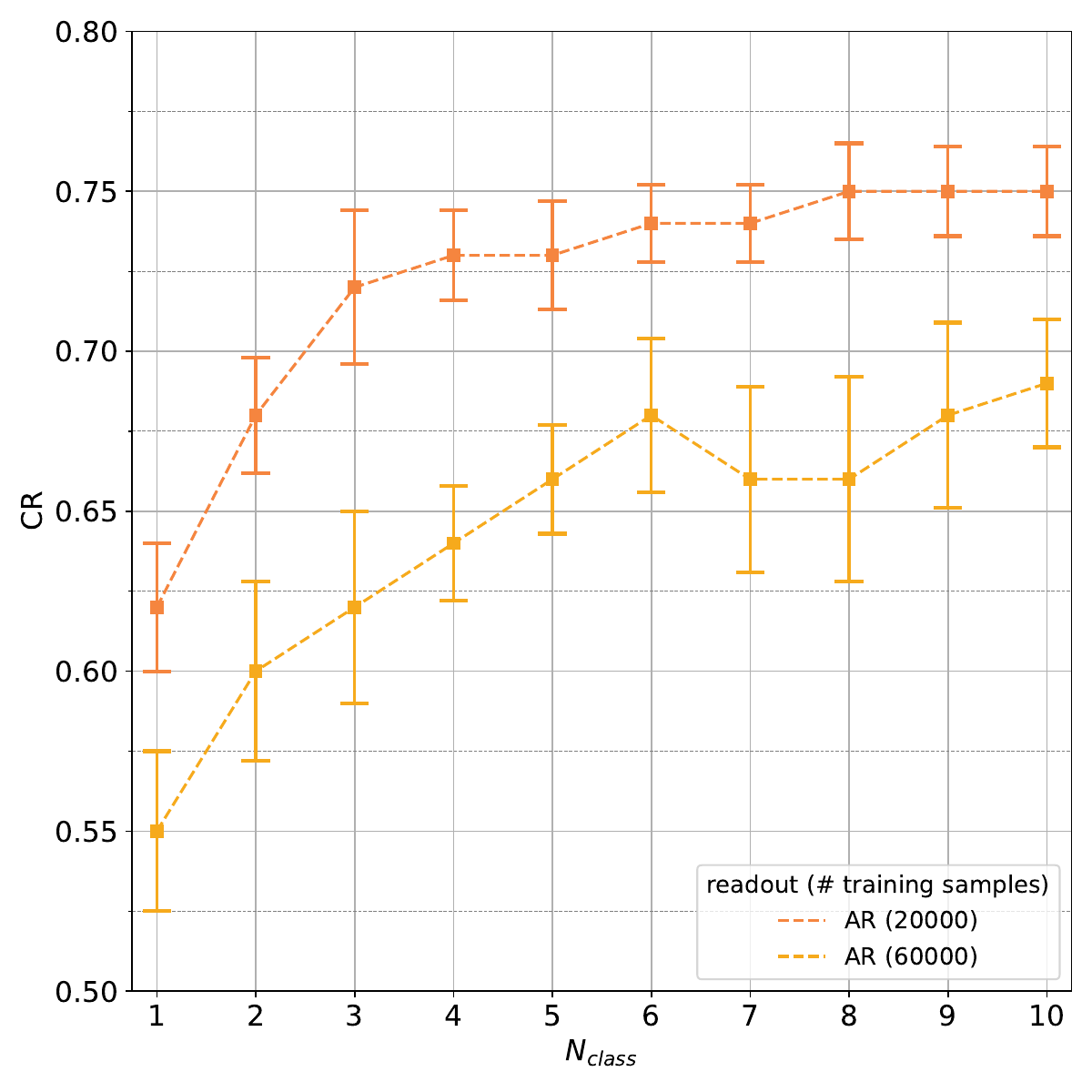}}
\caption{Comparison of network performance as a function of output pool size \gls{ar} voting method between networks training with 60k and 20k training samples. Error bars show $\pm1$ standard deviation. \gls{bcall} and \gls{sfnn} hyperparameters in tables \ref{tab:hyperparams_table} and \ref{tab:sfnn_variables}, respectively.}
\label{poolavg_60k_x_20K}
\end{figure}

We compare in \figurename~\ref{poolmax_60k_x_20K} and \ref{poolavg_60k_x_20K} the performance of the networks trained with reduced number of training samples versus the ones trained with the entire dataset. Although the scaling in performance is roughly the same, the \gls{cr} values obtained by networks trained with 20k samples are on average approximately $14\%$ and $10\%$ higher for the \gls{mr} and \gls{ar} readout methods, respectively. 

\begin{figure}[htbp]
\centerline{\includegraphics[width=10cm]{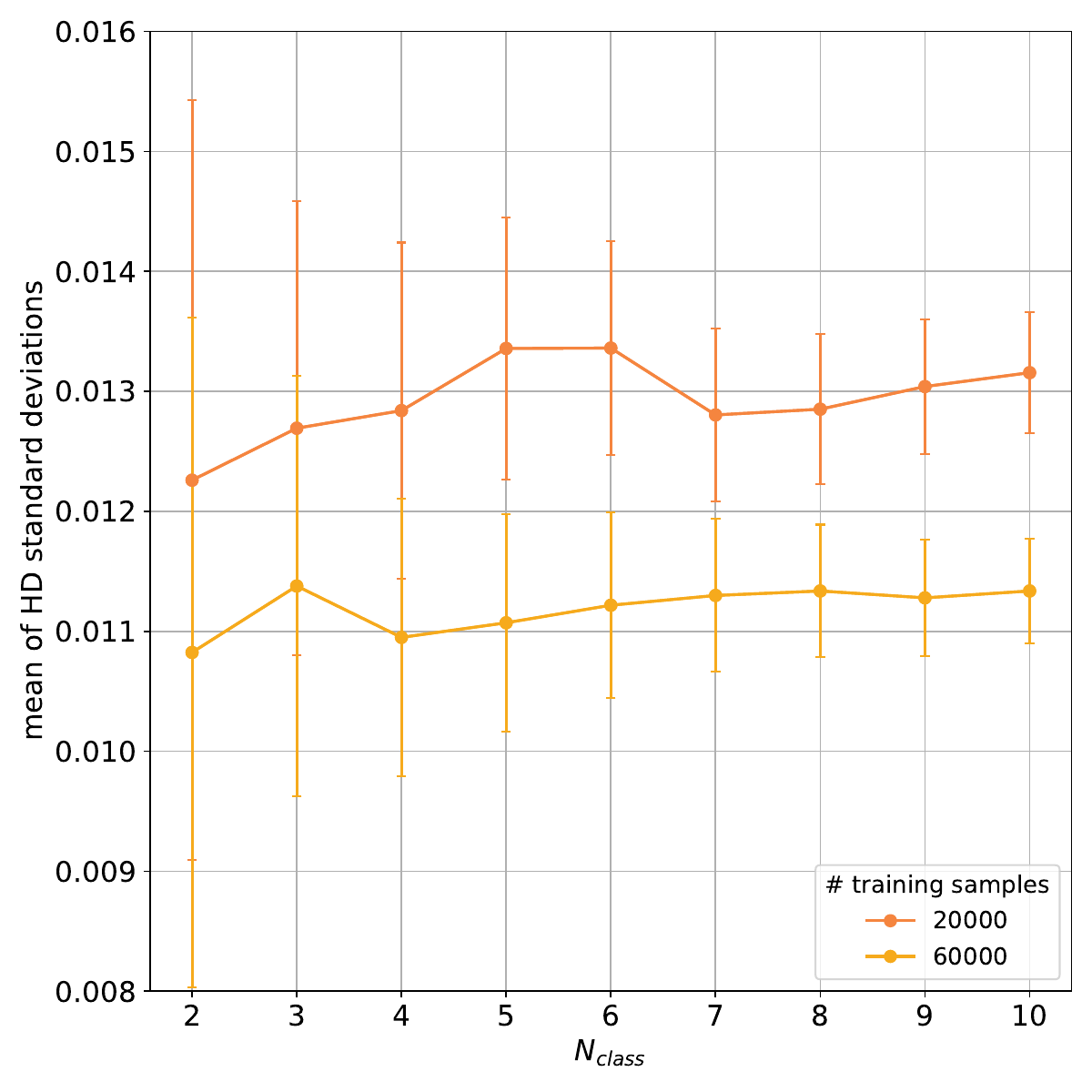}}
\caption{Comparison of mean standard deviation for the Hamming distances calculated between combinations of neurons within the same pools for all classes as a function of output pool size. Given that the number of possible combinations of matrices used to compute the \glspl{hd} increase quadratically with $N_{class}$, we show the error bar as the metric in the y-axis divided by the square root of the number of 2-by-2 matrices comparisons used to compute it.}
\label{poolvar_60k_x_20K}
\end{figure}

To check whether the increase in performance correlates with a higher variability inside each pool, we compute a standard deviation measure for the weight matrices' \glspl{hd} of neurons within the output pools. This is done by, for each pool (i.e., a set of output neurons encoding the same class), computing the \glspl{hd} between all combinations (without repetitions) of weight matrices, from which we compute the mean and standard deviation of these \glspl{hd} over all classes. Since we have trained ten independent \glspl{sfnn} for each pool size explored, the standard deviations computed for each pool in each of them is averaged. More formally described, every \gls{sfnn} gives us mean and standard deviations $\overline{X_1}, ..., \overline{X_{10}}$ and $var(X_1), ..., var(X_{10})$, respectively. We then compute the mean and standard deviation as $\overline{Y}=\frac{1}{10}\sum_{i=1}^{i=10}\overline{X_i}$ and $var(Y)=\frac{1}{10}\sum_{i=1}^{i=10}var(X_i)$, the latter being the measure shown in \figurename~\ref{poolvar_60k_x_20K} for each value of $N_{class}$. As it can be seen, when training with a reduced number of samples the within pool variability is increased (by approximately $18\%$), which translates into a higher \gls{cr} since the intra-class variability of samples can be better captured by the weight matrices learned by the output pools. 

The networks trained on the entire dataset tend to ``overfit'' the data: the repeated exposure to numerous samples of each class over time leads the weight matrices to converge toward a more uniform, idealized representation that inadequately captures the intra-class variability in the dataset. When the training dataset is reduced, the weight matrices retain more of the initial randomness from their initialization, as they have less exposure to samples that would otherwise refine them toward the average representation of a class. 

The scaling in performance with $N_{class}$ can be attributed to two factors: (1) the probabilistic nature of the weight updates and (2) the randomness of the initial weight initialization. A larger pool size introduces more variability in the weight matrices due to random initialization of their weights, resulting in slightly different matrices configurations. This is the upward trends with see in Fig.~\ref{poolvar_60k_x_20K}. That is, the performance increase is primarily driven by slight variations in random weight initialization, rather than neurons effectively learning complementary features (which would require some form of competition among neurons in the output pools). This explanation is consistent with the performance scaling saturation we observe.

\begin{figure}[htbp]
\centerline{\includegraphics[width=10cm]{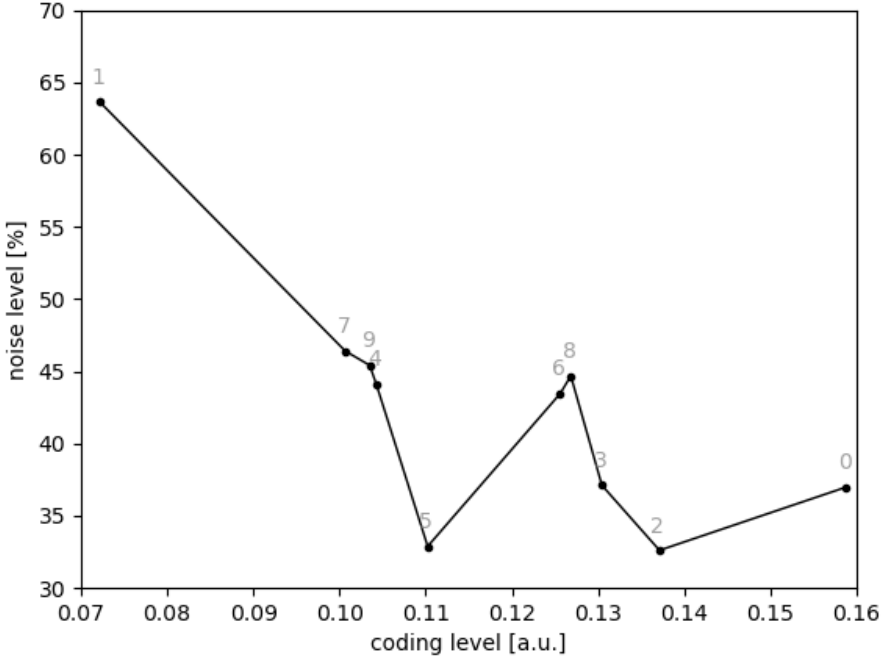}}
\caption{Percentage of noise within synaptic matrix as a function of the coding-level for the learned class. Values over data points indicate class label. Noise computed for a \gls{sfnn} trained on the entire MNIST dataset.}
\label{fig_9}
\end{figure}

In \figurename~\ref{fig_9} we show the percentage of noise in the learned representations as a function of the average coding-level of the encoded class. A weight is considered ``noise'' in such matrices if it does not match a pixel in the prototype (average digit image) of a class. As expected, classes with very low coding-level like class 1 require a higher percentage of noise in the synaptic matrix to compensate for it. Each of these ``noise synapses'' provides the post-synaptic neuron with a low rate synaptic input, which adds to the highly active pre-synaptic neurons representing features of the pattern and mitigates the a lower coding-level.

\section{Recurrent Spiking Neural Network}

Despite intense debate, whether \gls{stdp} is more fundamental than frequency-dependent plasticity rules is a debate that remains unsolved~\cite{Lisman_etal2005}. Both mechanisms play crucial roles in synaptic modification, yet they function based on different principles. More over, since \gls{stdp} depends on stimulation frequency, an interaction between timing- and frequency-dependent processes may be at play~\cite{Sjostrom_etal2001}. This ongoing debate suggests these mechanisms may be complementary, rather than hierarchical, in their roles within neural computation.

We aim to demonstrate one potential interplay between these two mechanisms using \gls{bcall}. Our learning rule, as shown, is capable of exhibiting both \gls{stdp} and \gls{srdp} plasticity outcomes without altering its hyperparameters. We hypothesize that timing can be leveraged to modulate the rate at which synapses are modified. Specifically, without changing the mean firing rate of neurons in the network nor any hyperparameters of the learning rule, the learning process can be accelerated by introducing correlations in their spike trains.

We chose to explore our hypothesis using a \gls{rsnn} architecture, given they are known for their ability to exhibit synchronized neuronal activity and are found in various brain regions. This synchronization often relies on rhythmic subthreshold fluctuations in membrane potentials~\cite{Chiang_etal2022}, such as those observed in oscillatory brain activity (e.g., theta, gamma rhythms). These oscillations are commonly observed in regions such as the hippocampus, cortex, and thalamus. In the hippocampus, subthreshold oscillations are thought to play a role in synchronizing neuronal activity and shaping the timing of spike outputs, which is crucial for processes such as memory encoding and retrieval~\cite{OKeefe_etal1993, Buzsaki_2002}. Subthreshold oscillations are also noted in the thalamus, where they are believed to influence the rhythmic bursting of thalamocortical neurons and their interactions with cortical networks~\cite{Neske_2016}.

\begin{figure}[t]
\centerline{\includegraphics[width=12cm]{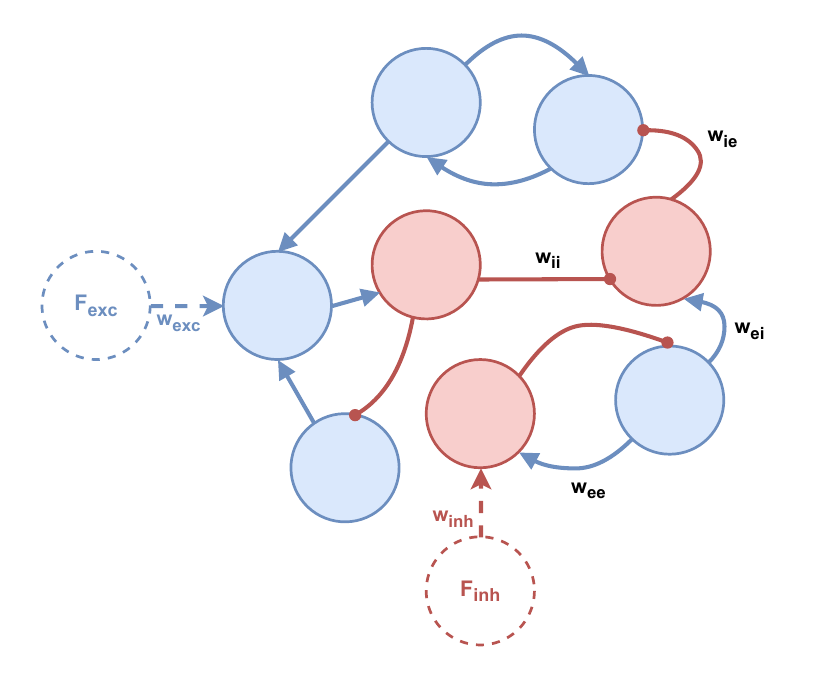}}
\caption{The recurrent network is composed of two main populations of neurons, a pool of 256 excitatory neurons (blue circles) and a pool of 64 inhibitory (red circles) neurons. Each neuron in the network receives Poisson spikes trains from ``virtual'' units (dashed circles) whose sole purpose is to drive the activity in the network. Excitatory synapses are represented as arrows and inhibitory ones as circles. The neuron model parameters are the same as the ones listed in Table~\ref{tab:neurons_table}, with exception of the variable $V_{incr}$, which is set to \SI{0}{\milli\volt} in order to keep the threshold adaption from interfering with the subthreshold membrane oscillations. \gls{bcall} hyperparameters used are the ones in Table~\ref{tab:hyperparams_table}, with the stop-learning mechanism being effectively disabled since it does not play a role in this task. The architectural parameters for the network are listed in Table~\ref{tab:recurrent_net_architecture}.}
\label{B_RCN}
\end{figure}

The \gls{rsnn} we use for our simulations is shown in \figurename~\ref{B_RCN}. The network is composed of excitatory and inhibitory \gls{lif} neurons with recurrent connections amongst them, without self-connections (i.e., a neuron does not synapse onto itself). The only plastic synapses in the network are the ones between the excitatory units ($w_{ee}$), while all remaining connections ($w_{ei}$, $w_{ie}$ and $w_{ii}$) are fixed. Similarly to our previously described \gls{sfnn}, each excitatoty and inhibitory neuron receive Poisson spike trains from ``virtual'' units whose sole purpose is to provide the input to drive the neurons in the network. The underlying connectivity pattern is randomly drawn and the connection probabilities between pairs, fixed weights' strength and number of excitatory and inhibitory units being taken from~\cite{Jug_etal2012}.

\begin{table}
    \centering
    \begin{tabular}{c p{0.35\linewidth} c}
 Hyperparameter & Description & Value\\
 \midrule
         $w_{ee}$ & Plastic connection weight between excitatory neurons. & \SI{3}{\milli\volt}\\
         $w_{ei}$ & Fixed connection weight from excitatory to inhibitory neurons. & \SI{3}{\milli\volt}\\
         $w_{ie}$ & Fixed connection weight from inhibitory to excitatory neurons. & \SI{2}{\milli\volt}\\
         $w_{ii}$ & Fixed connection weight between inhibitory neurons. & \SI{2}{\milli\volt}\\
         $w_{exc}$ & Fixed connection weight from virtual excitatory neurons. & \SI{1}{\milli\volt}\\
         $w_{inh}$ & Fixed connection weight from virtual inhibitory neurons. & \SI{10}{\milli\volt}\\
         $F_{exc}$ & Mean firing rate of virtual excitatory neurons. & \SI{1000}{\hertz}\\
         $F_{inh}$ & Mean firing rate of virtual inhibitory neurons. & \SI{200}{\hertz}\\
         $p_{ee}$ & Connection probability between excitatory neurons. & 0.5\\
         $p_{ei}$ & Connection probability from excitatory to inhibitory neurons. & 0.25\\
         $p_{ie}$ & Connection probability from inhibitory to excitatory neurons. & 0.25\\
         $p_{ii}$ & Connection probability between inhibitory neurons. & 0.5\\
         $f_{osc}$ & Subthreshold oscillation frequency. & \SI{3}{\hertz}\\
         $\phi_{rand}$ & Random subthreshold oscillation phase shift range. & [$-\pi$, $\pi$] radians\\
         $\phi_{corr}$ & Correlated subthreshold oscillation phase shift range. & [-0.1, 0.1] radians\\
 \midrule
    \end{tabular}
\caption{Hyperparameters used for the \gls{rsnn}.}
\label{tab:recurrent_net_architecture}
\end{table}

In such \gls{rsnn}, if a large enough subset of the excitatory neurons are stimulated for long enough (i.e., receive concurrent input for a period of time), the recurrent excitatory weights ($w_{ee}$) amongst them can be learned (i.e., go from the LOW to their UP state when $w_{hid} > \theta_{w}$) such that their activity can be maintained in the absence of the ``external'' input provided by the virtual units, forming what is commonly known as an attractor~\cite{Willian_Thomas2024}. Given the recurrency in the architecture we use here, the correlations in the spike trains provided by the virtual units would not necessarily translate into correlated spike trains in the excitatory neurons. Thus, in order to introduce such correlated activity, the excitatory units in the \gls{rsnn} also receive subthreshold sinusoidal inputs such that their spike probability is increased at the peak of the sinusoid. With this the membrane potential of such neurons is driven by a combination of synaptic inputs (both excitatory and inhibitory from within the network and the excitatory spike trains from the virtual units) and an input current that causes the oscillations. If the phase of the sinusoidal input is matched amongst the excitatory neurons, their spike activity becomes correlated, otherwise they are random. 

We test whether spiking correlations induced by subthreshold membrane oscillations can increase the learning rate of attractors by comparing the average $w_{hid}$ value of all synapses within a subset of 64 excitatory neurons that are driven to spike at the same mean firing rate. Simulations of the \glspl{rsnn} were performed with two different phase shift setups for the sinusoidal inputs: one where the phases were randomly sampled between $-\pi$ and $\pi$ radians, covering a full oscillation cycle (which we'll refer to as ``random''), and another where the phases were sampled from a narrow range between $-0.1$ and $0.1$ radians to induce closely aligned inputs with minimal variation (which we'll refer to as ``correlated''). A total of ten networks were simulated for each setup. In both cases the networks are simulated for one second. All plastic synapses are initialized to $w_{ee} = 0mV$, that is, all excitatory neurons are effectively disconnected and only influence each other once $w_{ee} > \theta_{w}$.

\begin{figure*}[t]
    \centering
    {{\includegraphics[width=\linewidth]{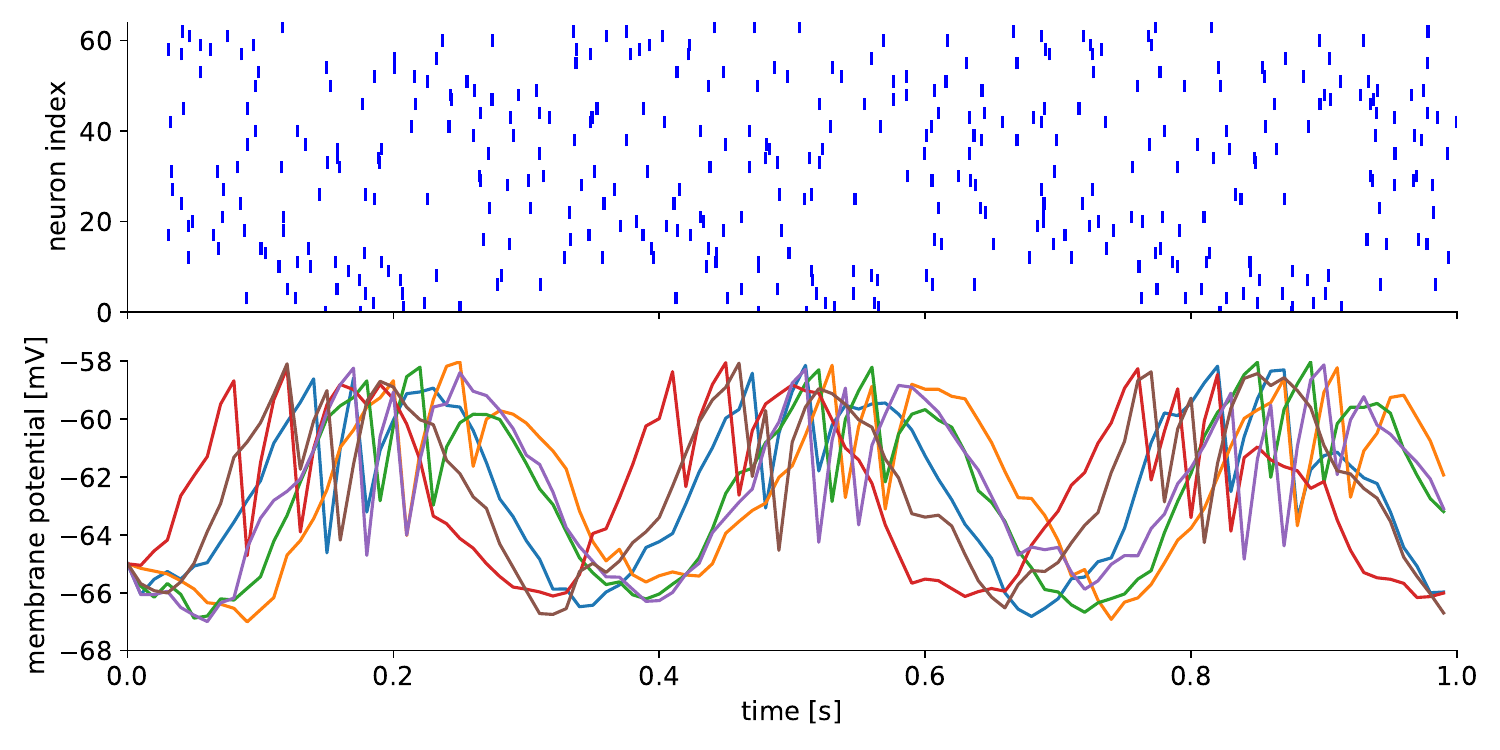}}}
    \caption{One second simulation of a recurrent network where the subthreshold membrane oscillations of excitatory neurons are randomly sampled between $-\pi$ and $\pi$ radians. The top plot shows the spiking activity of the subset of neurons that are being stimulated via Poisson spike trains over a period of one second. The bottom plots show the evolution over time of six randomly sampled membrane potentials, which shows that the neurons tend to fire at the peak of the sinusoid.}
    \label{rand_raster}
\end{figure*}

\begin{figure*}[t]
    \centering
    {{\includegraphics[width=\linewidth]{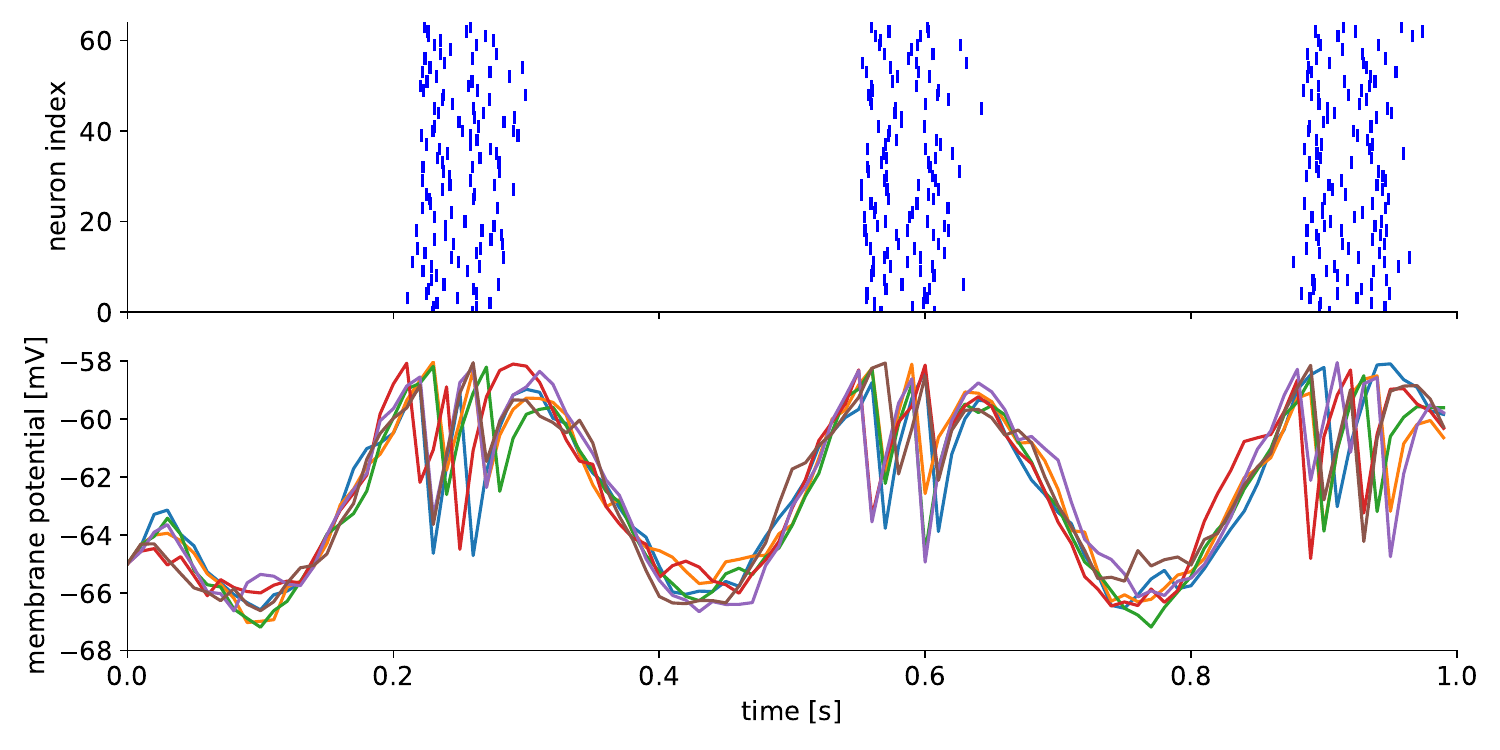}}}
    \caption{One second simulation of a recurrent network where the subthreshold membrane oscillations of excitatory neurons are randomly sampled between $-0.1$ and $0.1$ radians. The top plot shows the spiking activity of the subset of neurons that are being stimulated via Poisson spike trains over a period of one second. The bottom plots show the evolution over time of six randomly sampled membrane potentials: differently from what can be seen in \figurename~\ref{rand_raster}, the phases of the oscillations match within a very narrow range, leading to a more synchronized spiking activity.}
    \label{corr_raster}
\end{figure*}

We show what the activity in the simulated \glspl{rsnn} look like for the random and correlated oscillation phases setups in \figurename~\ref{rand_raster} and \figurename~\ref{corr_raster}, respectively. As expected, with the correlated setup the raster plot of the spiking activity in \figurename~\ref{corr_raster} (top row) show a much more closely timed overall spiking activity of the subset of neurons being stimulated to form an attractor due to the fact that the phases of their oscillatory input matches more narrowly (bottom row).

\begin{figure}[t]
    \centering
    {{\includegraphics[width=10cm]{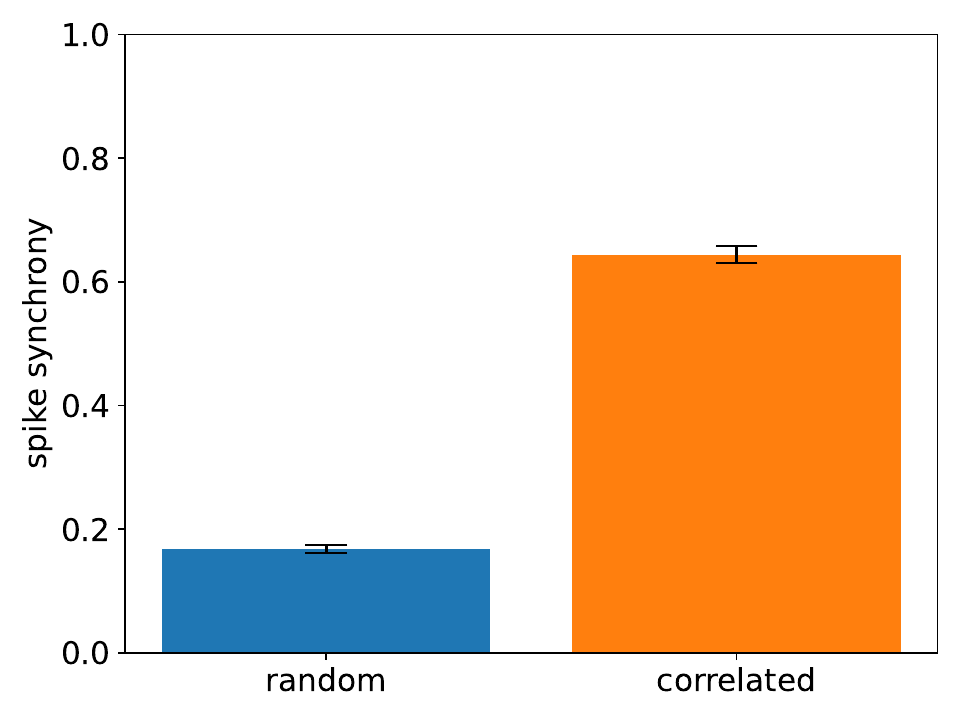}}}
    \caption{Degree of spike synchronization between the networks with random and correlated subthreshold membrane oscillation phases.}
    \label{spk_sync_comparison}
\end{figure}

The increased synchronization can be more formally visualized with the help of the SPIKE-synchronisation metric~\cite{Mulansky_etal2015}: this coincidence detection algorithm quantifies the degree of synchronisation in the spikes of subsets of neurons as a number between zero (uncorrelated firing) and one (perfect synchrony). We show the average SPIKE-synchronisation computed from the activity of the subset of excitatory neurons that were stimulated during the simulation of each of the ten networks under both setting in \figurename~\ref{spk_sync_comparison}. This data shows that the networks with correlated phases are approximately three times more synchronized than their random counterparts.

\begin{figure}
    \centering
    {{\includegraphics[width=10cm]{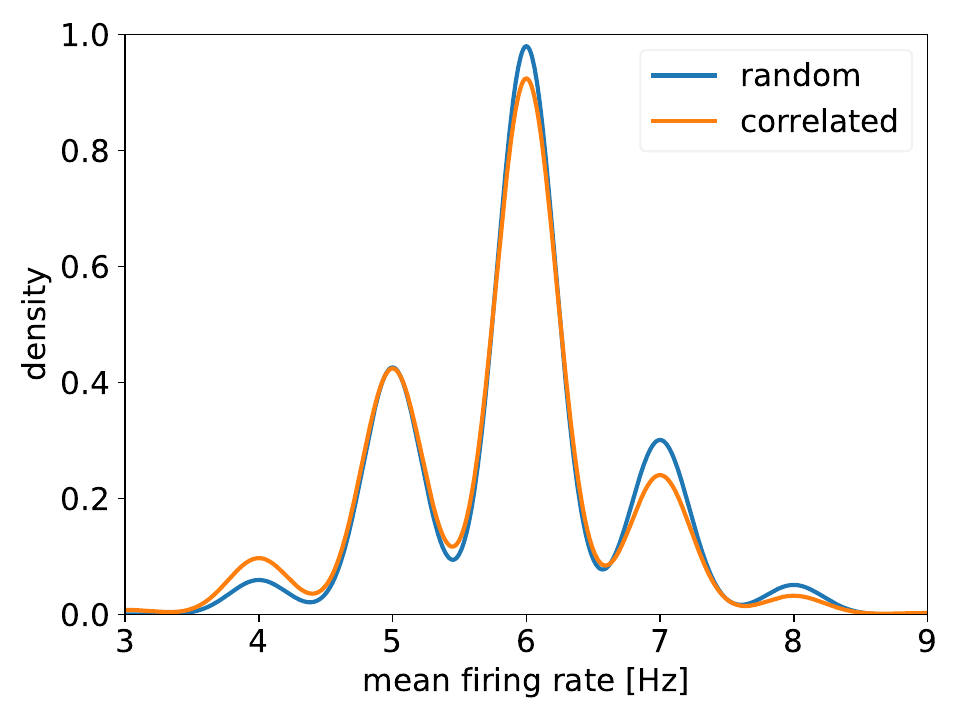}}}
    \caption{Mean firing rates histograms of all simulated networks for both sinusoidal oscillation setups. The mean firing rates of the subset of excitatory neurons that are stimulated during the simulation are virtually the same, showing that there are no differences in the mean rates that could impact the rate of change of the synapses.}
    \label{mean_firing_rate_comparison}
\end{figure}

In order to make sure that there are no differences in the mean firing rates of the neurons (due to the different phase setups) that could be influencing the evolution of the variables $w_{hid}$ during the period of stimulation, we gather the individual mean firing rates of each neuron in each of the ten simulated networks and compute their \glspl{pdf}, shown in \figurename~\ref{mean_firing_rate_comparison}. As it can be seen, the \glspl{pdf} are nearly identical, allowing us to confidently rule out any potential influence from variations in mean firing rates across these networks.

\begin{figure}
    \centering
    {{\includegraphics[width=10cm]{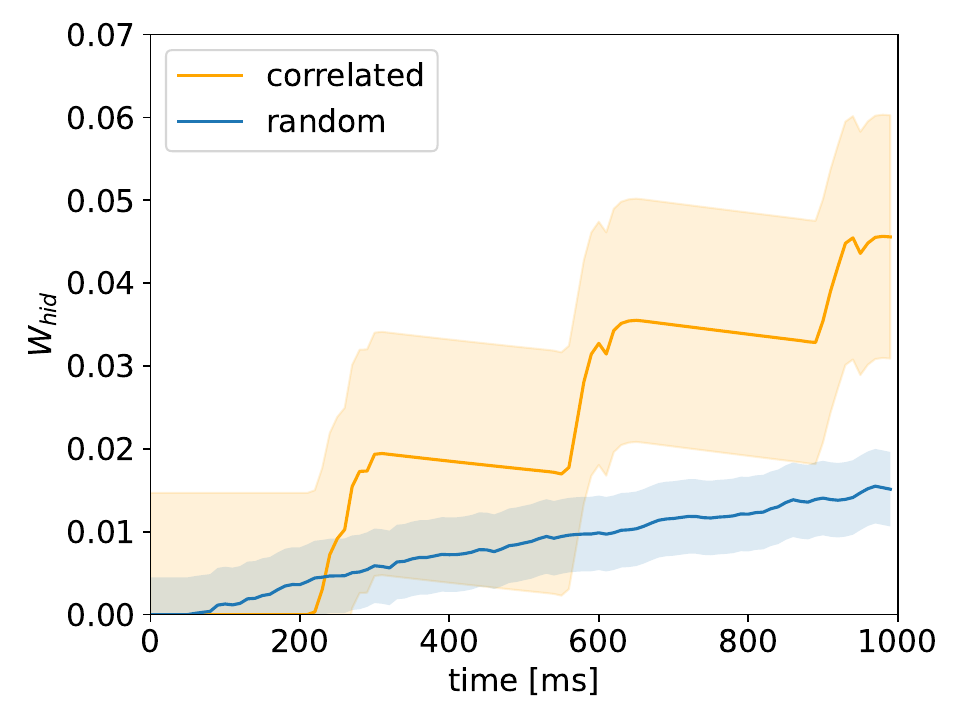}}}
    \caption{The plot shows a solid line representing the average $W_{hid}$ of ten simulated networks with different random connectivity, with a semi-transparent shaded region around the line indicating $\pm1$ standard deviation from the mean. Given that the mean firing rates are virtually identical across all networks in both phase setups, the differences seen in the average $W_{hid}$ at the end of the simulation are due to the increased synchrony. Thus, even though the mean firing rates are equal the learning rate is increased due to temporal correlation in the spiking activity of the neurons.}
    \label{mean_rho_comparison}
\end{figure}

Given that we have excluded any potential differences in the mean firing rates that could otherwise influence the evolution of the hidden weight state variable $w_{hid}$ amongst the subset of neurons that are being stimulated, any discrepancies seen for this variable at the end of the simulation should be due to correlations in the spike times of those neurons. To represent the evolution of this variable in a single simulation, we average the $w_{hid}$ associated with each synapse within the subset of neurons receiving external Poisson spiking input throughout the duration of the simulation, which will refer to as $W_{hid}^{i}$ for a single network instance $i$, with $1 \leq i \leq 10$. The solid lines in \figurename~\ref{mean_rho_comparison} represents the average $W_{hid}$ computed over all ten simulated networks for each setup of the phase shifts.

As shown in \figurename~\ref{mean_rho_comparison}, when neurons exhibit increased synchronicity (correlated phase shifts), the rate of increase of $W_{hid}^{i}$ is, on average, approximately four times higher compared to the case with uncorrelated spike times (random phase shifts). This illustrates the interplay between mean firing rate and spike timing within a network: while the mean firing rate establishes the baseline activity level of the neurons, spike timing serves as a ``knob'' that can be adjusted to modulate the learning rate within the system, without the need for modifying the hyperparameters of the learning rule nor the overall network level of activity.

\section{Discussion}

In this work we presented \gls{bcall}, a Hebbian learning rule designed as a mechanistic model of synaptic plasticity mediated by pre- and post-synaptic calcium transients. We show that the rule is able to qualitatively reproduce synaptic dynamics from well established \gls{stdp} and \gls{srdp} protocols. 

We validated the applicability of the learning rule to supervised learning by training single layer \gls{sfnn} for digits classification. While previous works have shown the necessity for a stop-learning condition and coding-level-dependent inhibition to properly train such networks, the focus of our analysis was to show in details \textit{why} such ingredients are necessary to train these networks on real-world patterns, \textit{what} roles they play and \textit{how} they interact to allow proper training. 

Contrary to what has been suggested before~\cite{Illing_etal2019}, we find that allowing the interaction of transient activity elicited during the switching of the network's input is beneficial for training since it minimizes the overlap between the learned representations. This could be a benefit of tracking spiking activity via traces: while some models might rely on the neuron's membrane voltage to compute weight updates, utilizing traces allows us to decouple from the membrane dynamics the extent to which correlated activity in time can be measured. Elucidating how computation in such systems can benefit from encoding information in such traces is not only important for being better able to understand the functional role of the complex biochemical machinery existing in biological neural networks but also for making better design decisions when building analog \glspl{asic} implementing these algorithms.

Our simulation results show that the increase in performance that comes with increasing the \glspl{sfnn} output layer is limited by how much diversity in a class representation additional units can create. This observation highlights the importance of diversification within class representations for achieving enhanced performance in single-layer spiking neural networks: by allowing for multiple distinct weight matrix representations of the same class, similarly to what is done with ensemble methods, the network becomes more resilient to variations and uncertainties in input patterns, leading to improved accuracy in classification tasks. It is essential to consider the potential limitations and drawbacks when such diversity cannot be guaranteed. In scenarios where networks lack this diversity in the representations despite increasing the output layer size, the benefits of enlarging the output layer become significantly diminished. Thus, allocating additional resources to a larger output layer without mechanisms that ensure that this diversity can be achieved incurs higher computational costs and increased memory requirements with diminishing returns in terms of performance improvement. 

A potential ``cheap'' strategy could be to leverage nonidealities (i.e., mismatch) in neuromorphic analog hardware to increase the variance within the matrix representations of the classes learned (synaptic-level mismatch) and neuronal responses (neuron-level mismatch).

We propose that future research could explore a bio-inspired mechanism to enhance diversity in the representations within an output pool encoding for a single class in the \gls{sfnn} architecture we used. Specifically, incorporating subthreshold sinusoidal oscillations in the input and output layer neurons could be effective. By assigning discrete phase shifts to various regions of the input space, output neurons within the same pool could adopt different phase shifts corresponding to those regions. This approach may bias individual output neurons toward recognizing distinct areas of the input space, resulting in more diverse synaptic matrices that capture a wider range of features.

As reported by Brader and colleagues~\cite{Brader_etal2007}, we also find that reducing the number of MNIST samples used to train the \gls{sfnn} improves accuracy. This enhancement arises from increased within-pool variability with fewer samples. When using the complete training dataset (60k samples), the learned weight matrices tend to converge toward an ``average'' representation of the digit, as output neurons within the same pool are exposed to multiple samples of the same class. In contrast, fewer samples increase the likelihood that different neurons will capture various features of the input space. This suggests that adjusting the hyperparameters of the learning rule—both in our work and in Brader et al.'s study—can facilitate slower feature learning.

The differences we show in the readout methods during inference have important implications when dealing with physical implementations of learning systems such as the one we explored here: given the complexity limitations arising not only from the learning paradigm and the simple \gls{sfnn} architecture but also from the physical limitations that \glspl{asic} with constrained resources impose, network performance can be improved by careful consideration on how to read out network activity.

Our work has demonstrated that with the \gls{bcall} rule the rate of change of synaptic couplings between the neurons in a \gls{rsnn} can be modulated via the incorporation of correlations in their spike times. Such correlations are introduced via subthreshold sinusoisal membrane potential oscillations in a way that mimics what is observed in some brain regions. We effectively show that the spike timing component of the rule can be leveraged to act as a ``knob'' to change the speed with which the weights in the network evolve, without making any alterations to the rule's hyperparameters nor the mean firing rate of neurons. While other works~\cite{Pfister_Gerstner2006, Clopath_etal2010} have shown that similar learning rules can reproduce both \gls{stdp} and \gls{srdp} plasticity outcomes, they fall short of showing how both mean firing rate and spike timing can coexist within a network and be leveraged at the same time. Here we show that in fact these mechanisms can be complementary in their roles within neural computation.

This finding is particularly relevant when simulating and studying learning in models of brain structures such as the \gls{ion}. In a paper~\cite{VanDerGiessen_etal2008} discussing the significance of electrical coupling in the \gls{ion}, researchers found that disrupting the electrical coupling between neurons had a detrimental effect on mice's ability to learn a motor task requiring coordinated movements across multiple joints due to lack of synchronization, which is vital for the formation of precise motor memories. The ability to modulate learning rates through synchronization, as we have shown, provides a valuable starting point for understanding this type of intricate learning dynamics.

The subthreshold oscillations used in our \glspl{rsnn} simulations, combined with the demonstrated modulation of the learning rate through spike-pair correlations, could prove particularly useful for training more complex recurrent network models. For instance, the phasic attractors model for context-dependent computation proposed by Soares Gir\~ao and Tiotto~\cite{Willian_Thomas2024} would require somewhat precisely timed attractor activation to learn the connections that enable state transitions within the model. This type of attractor network offers a compelling platform to further investigate the interplay between mean firing rate and spike timing, a focus we have initiated in this work.

\bibliographystyle{unsrt}  
\bibliography{references_BCaLL_new}  

\begin{thebibliography}{10}

\bibitem{Hebb_1999}
Morris RG.
\newblock D.{{O}}. hebb: {{The}} organization of behavior.
\newblock {\em Brain research bulletin}, 50(5-6):437, 1999.

\bibitem{Baudry_Lynch1979}
M.~Baudry and G.~Lynch.
\newblock Regulation of glutamate receptors by cations.
\newblock {\em Nature}, 282:748--750, 1979.

\bibitem{Lynch_etal1983}
G.~Lynch, J.~Larson, and S.~et~al. Kelso.
\newblock Intracellular injections of {{EGTA}} block induction of hippocampal
  long-term potentiation.
\newblock {\em Nature}, 305:719--721, 1983.

\bibitem{Turner_etal1982}
{\relax RW}~Turner, {\relax KG}~Baimbridge, and {\relax JJ}~Miller.
\newblock Calcium-induced long-term potentiation in the hippocampus.
\newblock {\em Neuroscience}, 7(6):1411--6, 1982.

\bibitem{Malenka_etal1988}
{\relax RC}~Malenka, {\relax JA}~Kauer, {\relax RS}~Zucker, and {\relax
  RA}~Nicoll.
\newblock Postsynaptic calcium is sufficient for potentiation of hippocampal
  synaptic transmission.
\newblock {\em Science (New York, N.Y.)}, 242(4875):81--4, 1988.

\bibitem{Ekerot_Kano1985}
{\relax CF}~Ekerot and M~Kano.
\newblock Long-term depression of parallel fibre synapses following stimulation
  of climbing fibres.
\newblock {\em Brain research}, 342(2):357--60, 1985.

\bibitem{Sakurai1990}
M~Sakurai.
\newblock Calcium is an intracellular mediator of the climbing fiber in
  induction of cerebellar long-term depression.
\newblock {\em Proceedings of the National Academy of Sciences of the United
  States of America}, 87(9):3383--5, 1990.

\bibitem{Brocher_etal1992}
S.~Br{\"o}cher, A.~Artola, and W.~Singer.
\newblock Intracellular injection of {{Ca2}}+ chelators blocks induction of
  long-term depression in rat visual cortex.
\newblock {\em Proceedings of the National Academy of Sciences of the United
  States of America}, 89(1):123--7, 1992.

\bibitem{Pfister_Gerstner2006}
{\relax JP}~Pfister and W~Gerstner.
\newblock Triplets of spikes in a model of spike timing-dependent plasticity.
\newblock {\em Journal of Neuroscience}, 26:9673--9682, 2006.

\bibitem{Fusi_etal2000}
S~Fusi, M~Annunziato, D~Badoni, A~Salamon, and {\relax DJ}~Amit.
\newblock Spike-driven synaptic plasticity: Theory, simulation, {{VLSI}}
  implementation.
\newblock {\em Neural computation}, 12(10):2227--58, 2000.

\bibitem{Diehl_Cook2015}
{\relax PU}~Diehl and M~Cook.
\newblock Unsupervised learning of digit recognition using
  spike-timing-dependent plasticity.
\newblock {\em Frontiers in computational neuroscience}, 9:99, 2015.

\bibitem{Kheradpisheh_etal2018}
Saeed~Reza Kheradpisheh, Mohammad Ganjtabesh, Simon~J. Thorpe, and Timoth{\'e}e
  Masquelier.
\newblock {{STDP-based}} spiking deep convolutional neural networks for object
  recognition.
\newblock {\em Neural Networks}, 99:56--67, 2018.

\bibitem{Zucker_Regehr2002}
{\relax RS}~Zucker and {\relax WG}~Regehr.
\newblock Short-term synaptic plasticity.
\newblock {\em Annual review of physiology}, 64:355--405, 2002.

\bibitem{Citri_Malenka2008}
A~Citri and R~Malenka.
\newblock Synaptic plasticity: {{Multiple}} forms, functions, and mechanisms.
\newblock {\em Neuropsychopharmacology}, 33:18--41, 2008.

\bibitem{Yang_Calakos2013}
Y~Yang and N~Calakos.
\newblock Presynaptic long-term plasticity.
\newblock {\em Frontiers in synaptic neuroscience}, 5:8, 2013.

\bibitem{Wittenberg_Wang2006}
{\relax GM}~Wittenberg and {\relax SS}~Wang.
\newblock Malleability of spike-timing-dependent plasticity at the {{CA3-CA1}}
  synapse.
\newblock {\em The Journal of neuroscience : the official journal of the
  Society for Neuroscience}, 26(24):6610--7, 2006.

\bibitem{Bi_Poo1998}
{\relax GQ}~Bi and {\relax MM}~Poo.
\newblock Synaptic modifications in cultured hippocampal neurons: Dependence on
  spike timing, synaptic strength, and postsynaptic cell type.
\newblock {\em The Journal of neuroscience : the official journal of the
  Society for Neuroscience}, 18(24):10464--72, 1998.

\bibitem{Sjostrom_etal2001}
{\relax PJ}~Sj{\"o}str{\"o}m, {\relax GG}~Turrigiano, and {\relax SB}~Nelson.
\newblock Rate, timing, and cooperativity jointly determine cortical synaptic
  plasticity.
\newblock {\em Neuron}, 32(6):1149--64, 2001.

\bibitem{Markram_etal1997}
H~Markram, J~L{\"u}bke, M~Frotscher, and B~Sakmann.
\newblock Regulation of synaptic efficacy by coincidence of postsynaptic
  {{APs}} and {{EPSPs}}.
\newblock {\em Science (New York, N.Y.)}, 275(5297):213--5, 1997.

\bibitem{Malenka_Bear2004}
Robert~C. Malenka and Mark~F. Bear.
\newblock {{LTP}} and {{LTD}}: {{An}} embarrassment of riches.
\newblock {\em Neuron}, 44(1):5--21, 2004.

\bibitem{Abbott_Nelson2000}
L.~Abbott and S.~Nelson.
\newblock Synaptic plasticity: Taming the beast.
\newblock {\em Nature neuroscience}, 3(Suppl 11):1178--1183, 2000.

\bibitem{Gulyaeva2017}
{\relax NV}~Gulyaeva.
\newblock Molecular mechanisms of neuroplasticity: {{An}} expanding universe.
\newblock {\em Biochemistry. Biokhimi\t{ia}}, 82(3):237--242, 2017.

\bibitem{Sjostrom_etal2008}
{\relax PJ}~Sj{\"o}str{\"o}m, {\relax EA}~Rancz, A~Roth, and M~H{\"a}usser.
\newblock Dendritic excitability and synaptic plasticity.
\newblock {\em Physiological reviews}, 88(2):769--840, 2008.

\bibitem{Brader_etal2007}
J.~Brader, W.~Senn, and S.~Fusi.
\newblock Learning real-world stimuli in a neural network with spike-driven
  synaptic dynamics.
\newblock {\em Neural Computation}, 19(11):2881--2912, 2007.

\bibitem{Graupner_Brunel2007}
M~Graupner and N~Brunel.
\newblock {{STDP}} in a bistable synapse model based on {{CaMKII}} and
  associated signaling pathways.
\newblock {\em PLoS computational biology}, 3(11):e221, 2007.

\bibitem{Graupner_Brunel2010}
M~Graupner and N~Brunel.
\newblock Mechanisms of induction and maintenance of spike-timing dependent
  plasticity in biophysical synapse models.
\newblock {\em Frontiers in computational neuroscience}, 4:136, 2010.

\bibitem{Graupner_Brunel2012}
M~Graupner and N~Brunel.
\newblock Calcium-based plasticity model explains sensitivity of synaptic
  changes to spike pattern, rate, and dendritic location.
\newblock {\em Proceedings of the National Academy of Sciences of the United
  States of America}, 109(10):3991--6, 2012.

\bibitem{Markram_Tsodyks1996}
H~Markram and M~Tsodyks.
\newblock Redistribution of synaptic efficacy between neocortical pyramidal
  neurons.
\newblock {\em Nature}, 382(6594):807--10, 1996.

\bibitem{Sjostrom_etal2003}
{\relax PJ}~Sj{\"o}str{\"o}m, {\relax GG}~Turrigiano, and {\relax SB}~Nelson.
\newblock Neocortical {{LTD}} via coincident activation of presynaptic {{NMDA}}
  and cannabinoid receptors.
\newblock {\em Neuron}, 39(4):641--54, 2003.

\bibitem{Lisman2003}
J~Lisman.
\newblock Long-term potentiation: Outstanding questions and attempted
  synthesis.
\newblock {\em Philosophical transactions of the Royal Society of London.
  Series B, Biological sciences}, 358(1432):829--42, 2003.

\bibitem{Sjostrom_etal2007}
{\relax PJ}~Sj{\"o}str{\"o}m, {\relax GG}~Turrigiano, and {\relax SB}~Nelson.
\newblock Multiple forms of long-term plasticity at unitary neocortical layer 5
  synapses.
\newblock {\em Neuropharmacology}, 52(1):176--84, 2007.

\bibitem{Morrison_etal2008}
A~Morrison, M~Diesmann, and W~Gerstner.
\newblock Phenomenological models of synaptic plasticity based on spike timing.
\newblock {\em Biological cybernetics}, 98(6):459--78, 2008.

\bibitem{Khacef_etal2023}
Lyes Khacef, Philipp Klein, Matteo Cartiglia, Arianna Rubino, Giacomo Indiveri,
  and Elisabetta Chicca.
\newblock Spike-based local synaptic plasticity: A survey of computational
  models and neuromorphic circuits.
\newblock {\em Neuromorphic Computing and Engineering}, 2023.

\bibitem{Willian_etal2024_cognigr1}
Willian Soares~Girão, Ole Richter, Philipp Klein, Michele Mastella, Hugh
  Greatorex, Madison Cotteret, Ella Janotte, and Elisabetta Chicca.
\newblock A subthreshold cmos hardware implementation of a bistable
  calcium-based local learning rule.
\newblock Manuscript in preparation, 2024.

\bibitem{Petersen_etal1998}
{\relax CC}~Petersen, {\relax RC}~Malenka, {\relax RA}~Nicoll, and {\relax
  JJ}~Hopfield.
\newblock All-or-none potentiation at {{CA3-CA1}} synapses.
\newblock {\em Proceedings of the National Academy of Sciences of the United
  States of America}, 95(8):4732--7, 1998.

\bibitem{OConnor_etal2005}
{\relax DH}~O'Connor, {\relax GM}~Wittenberg, and {\relax SS}~Wang.
\newblock Graded bidirectional synaptic plasticity is composed of switch-like
  unitary events.
\newblock {\em Proceedings of the National Academy of Sciences of the United
  States of America}, 102(27):9679--84, 2005.

\bibitem{sonntag_2020}
Annkathrin Sonntag.
\newblock Synaptic plasticity in spiking neural networks.
\newblock Master's thesis, Bielefeld University, Bielefeld, Germany, 2020.

\bibitem{Dayan_Abbott2005}
Peter Dayan and Laurence~F Abbott.
\newblock {\em Theoretical neuroscience: computational and mathematical
  modeling of neural systems}.
\newblock MIT press, 2005.

\bibitem{LeCun_etal}
Yann LeCun, Corinna Cortes, and Christopher~J.C. Burges.
\newblock {{THE MNIST DATABASE}} of handwritten digits.

\bibitem{Stimberg2019}
Marcel Stimberg, Romain Brette, and Dan~FM Goodman.
\newblock Brian 2, an intuitive and efficient neural simulator.
\newblock {\em eLife}, 8:e47314, August 2019.

\bibitem{Illing_etal2019}
Bernd Illing, Wulfram Gerstner, and Johanni Brea.
\newblock Biologically plausible deep learning \textemdash{} {{But}} how far
  can we go with shallow networks?
\newblock {\em Neural Networks}, 118:90--101, 2019.

\bibitem{Lisman_etal2005}
John Lisman and Nelson Spruston.
\newblock Postsynaptic depolarization requirements for {LTP} and {LTD}: a
  critique of spike timing-dependent plasticity.
\newblock {\em Nat Neurosci}, 8(7):839--841, July 2005.

\bibitem{Chiang_etal2022}
Chia-Chu Chiang and Dominique~M Durand.
\newblock Subthreshold oscillating waves in neural tissue propagate by volume
  conduction and generate interference.
\newblock {\em Brain Sci}, 13(1), December 2022.

\bibitem{OKeefe_etal1993}
J~O'Keefe and M~L Recce.
\newblock Phase relationship between hippocampal place units and the {EEG}
  theta rhythm.
\newblock {\em Hippocampus}, 3(3):317--330, July 1993.

\bibitem{Buzsaki_2002}
Gy{\"o}rgy Buzs{\'a}ki.
\newblock Theta oscillations in the hippocampus.
\newblock {\em Neuron}, 33(3):325--340, January 2002.

\bibitem{Neske_2016}
Garrett~T. Neske.
\newblock The slow oscillation in cortical and thalamic networks: Mechanisms
  and functions.
\newblock {\em Frontiers in Neural Circuits}, 9, 2016.

\bibitem{Jug_etal2012}
Florian Jug, Matthew Cook, and Angelika Steger.
\newblock {Recurrent Competitive Networks Can Learn Locally Excitatory
  Topologies}.
\newblock {\em The 2012 International Joint Conference on Neural Networks
  (IJCNN)}, 1:1--8, 2012.

\bibitem{Willian_Thomas2024}
Willian Soares~Girão and Thomas Tiotto.
\newblock Phasic attractors for flexible and adaptive working memory in spiking
  recurrent neural networks.
\newblock Manuscript under review at IOPscience, July 2024.

\bibitem{Mulansky_etal2015}
Mario Mulansky, Nebojsa Bozanic, Andreea Sburlea, and Thomas Kreuz.
\newblock A guide to time-resolved and parameter-free measures of spike train
  synchrony.
\newblock In {\em 2015 International Conference on Event-Based Control,
  Communication, and Signal Processing ({{EBCCSP}})}. {IEEE}, 2015.

\bibitem{Clopath_etal2010}
Claudia Clopath, Lars Büsing, Eleni Vasilaki, and Wulfram Gerstner.
\newblock {Connectivity reflects coding: a model of voltage-based STDP with
  homeostasis}.
\newblock {\em Nature Neuroscience}, 13(3):344--352, 2010.

\bibitem{VanDerGiessen_etal2008}
{\relax RS}~Van Der~Giessen, {\relax SK}~Koekkoek, S~{van Dorp}, {\relax
  JR}~De~Gruijl, A~Cupido, S~Khosrovani, B~Dortland, K~Wellershaus, J~Degen,
  J~Deuchars, {\relax EC}~Fuchs, H~Monyer, K~Willecke, {\relax MT}~De~Jeu, and
  {\relax CI}~De~Zeeuw.
\newblock Role of olivary electrical coupling in cerebellar motor learning.
\newblock {\em Neuron}, 58(4):599--612, 2008.

\end{thebibliography}

\end{document}